\documentclass{article}
 
\usepackage[final]{nips_2018}
\usepackage[utf8]{inputenc} 
\usepackage[T1]{fontenc}
 \usepackage{url}            
\usepackage{booktabs} 
\usepackage{amsfonts}       
\usepackage{nicefrac}       
\usepackage{microtype}
\usepackage[destlabel]{hyperref}
\usepackage{enumitem}
\setlist[enumerate,1]{leftmargin=2em,labelindent=0em,itemsep=0pt,labelsep*=0.5em}
\setlist[itemize,1]{leftmargin=2em,labelindent=0em,itemsep=0pt,labelsep*=0.5em}
\usepackage{wrapfig}

\author{
  Ji Xu \\
  Columbia University\\
  \texttt{jixu@cs.columbia.edu} 
  \And
  Daniel Hsu \\
  Columbia University\\
  \texttt{djhsu@cs.columbia.edu} 
  \And
  Arian Maleki \\
  Columbia University\\
  \texttt{arian@stat.columbia.edu} 
} 
 
  \usepackage{color}
  \usepackage{amsmath,amsthm}
  \usepackage{mathrsfs}
  \usepackage{extarrows}
  \usepackage{amssymb}
  \usepackage{graphicx,nicefrac}
\usepackage{makecell,multirow,diagbox}

  \newtheorem{theorem}{Theorem}
  \newtheorem{lemma}{Lemma}
  \newtheorem{corollary}{Corollary}
  \newtheorem{remark}{Remark}
  
  \definecolor{purple}{rgb}{0.8, 0.1, 0.5}

\newcommand\vpt[1]{\boldsymbol{\theta}^{\langle #1 \rangle}}  
\newcommand\pt[1]{\theta^{\langle #1 \rangle}}  
\newcommand\pw[1]{w^{\langle #1 \rangle}}   
\newcommand\pa[1]{\alpha^{\langle #1 \rangle}}   
\newcommand\pb[1]{\beta^{\langle #1 \rangle}}   
\newcommand\ptpa[1]{\theta^{\langle #1 \rangle}_{\parallel}}   
\newcommand\ptpe[1]{\theta^{\langle #1 \rangle}_{\perp}}   
\newcommand\tpt[1]{\tilde{\theta}^{\langle #1 \rangle}}   

\newcommand\hvpt[1]{\hat{\boldsymbol{\theta}}^{\langle #1 \rangle}}  
\newcommand\hpw[1]{\hat{w}^{\langle #1 \rangle}}   

\def\taga{\theta_{\gamma}}

\def\n{\nonumber}
\def\kgeq{ \ \geq \ }
\def\kleq{ \ \leq \ }
\def\={\ = \ }
\def\<{\ < \ }
\def\>{\ > \ }
\def\kand{\quad \text{and} \quad }
\usepackage{amsmath,amsbsy,amsfonts,amssymb,amsthm,dsfont,mleftright,commath}

\def\ddefloop#1{\ifx\ddefloop#1\else\ddef{#1}\expandafter\ddefloop\fi}

\def\ddef#1{\expandafter\def\csname bf#1\endcsname{\ensuremath{\mathbf{#1}}}}
\ddefloop ABCDEFGHIJKLMNOPQRSTUVWXYZabcdefghijklmnopqrstuvwxyz\ddefloop

\def\ddef#1{\expandafter\def\csname bf#1\endcsname{\ensuremath{\pmb{\csname #1\endcsname}}}}
\ddefloop {alpha}{beta}{gamma}{delta}{epsilon}{varepsilon}{zeta}{eta}{theta}{vartheta}{iota}{kappa}{lambda}{mu}{nu}{xi}{pi}{varpi}{rho}{varrho}{sigma}{varsigma}{tau}{upsilon}{phi}{varphi}{chi}{psi}{omega}{Gamma}{Delta}{Theta}{Lambda}{Xi}{Pi}{Sigma}{varSigma}{Upsilon}{Phi}{Psi}{Omega}{ell}\ddefloop

\def\ddef#1{\expandafter\def\csname bb#1\endcsname{\ensuremath{\mathbb{#1}}}}
\ddefloop ABCDEFGHIJKLMNOPQRSTUVWXYZ\ddefloop

\def\ddef#1{\expandafter\def\csname c#1\endcsname{\ensuremath{\mathcal{#1}}}}
\ddefloop ABCDEFGHIJKLMNOPQRSTUVWXYZ\ddefloop

\def\ddef#1{\expandafter\def\csname v#1\endcsname{\ensuremath{\boldsymbol{#1}}}}
\ddefloop ABCDEFGHIJKLMNOPQRSTUVWXYZabcdefghijklmnopqrstuvwxyz\ddefloop

\def\ddef#1{\expandafter\def\csname v#1\endcsname{\ensuremath{\boldsymbol{\csname #1\endcsname}}}}
\ddefloop {alpha}{beta}{gamma}{delta}{epsilon}{varepsilon}{zeta}{eta}{theta}{vartheta}{iota}{kappa}{lambda}{mu}{nu}{xi}{pi}{varpi}{rho}{varrho}{sigma}{varsigma}{tau}{upsilon}{phi}{varphi}{chi}{psi}{omega}{Gamma}{Delta}{Theta}{Lambda}{Xi}{Pi}{Sigma}{varSigma}{Upsilon}{Phi}{Psi}{Omega}{ell}\ddefloop

\renewcommand\t{{\ensuremath{\scriptscriptstyle{\top}}}}

\renewcommand\v{\ensuremath{\boldsymbol}}

\newcommand\dotp[1]{\langle #1 \rangle}

\newcommand\SBEM[1]{\ensuremath{\text{Sample-based~$\text{EM}_{#1}$}}}
\newcommand\PEM[1]{\ensuremath{\text{Population~$\text{EM}_{#1}$}}}

\newcommand\spacefrac[2]{#1\; / \;#2}

\title{Benefits of over-parameterization with EM}

\begin{document}
\maketitle
\begin{abstract}
Expectation Maximization (EM) is among the most popular algorithms for maximum likelihood estimation, but it is generally only guaranteed to find its stationary points of the log-likelihood objective. The goal of this article is to present theoretical and empirical evidence that over-parameterization can help EM avoid spurious local optima in the log-likelihood. We consider the problem of estimating the mean vectors of a Gaussian mixture model in a scenario where the mixing weights are known. Our study shows that the global behavior of EM, when one uses an over-parameterized model in which the mixing weights are treated as unknown, is better than that when one uses the (correct) model with the mixing weights fixed to the known values. For symmetric Gaussians mixtures with two components, we prove that introducing the (statistically redundant) weight parameters enables EM to find the global maximizer of the log-likelihood starting from almost any initial mean parameters, whereas EM without this over-parameterization may very often fail. For other Gaussian mixtures, we provide empirical evidence that shows similar behavior. Our results corroborate the value of over-parameterization in solving non-convex optimization problems, previously observed in other domains.

\end{abstract}
\section{Introduction}


In a Gaussian mixture model (GMM), the observed data $\mathcal{Y}=\{\vy_1,\vy_2,\ldots, \vy_n\} \subset \bbR^d$ comprise an i.i.d.~sample from a mixture of $k$ Gaussians:
\begin{eqnarray}
\vy_1,\dotsc,\vy_n&\stackrel{\text{i.i.d.}}{\sim}& \sum_{i=1}^k w^*_i \ \mathcal{N}(\vtheta_i^*, \vSigma^*_i) \label{eq:model1}
\end{eqnarray}
where $(w^*_i, \vtheta^*_i, \vSigma^*_i)$ denote the weight, mean, and covariance matrix of the $i^{\rm th}$ mixture component. Parameters of the GMM are often estimated using the Expectation Maximization (EM) algorithm, which aims to find the maximizer of the log-likelihood objective. However, the log-likelihood function is not concave, so EM is only guaranteed to find its stationary points. This leads to the following natural and fundamental question in the study of EM and non-convex optimization: How can EM escape spurious local maxima and saddle points to reach the maximum likelihood estimate (MLE)? In this work, we give theoretical and empirical evidence that over-parameterizing the mixture model can help EM achieve this objective.

Our evidence is based on models in~\eqref{eq:model1} where the mixture components share a known, common covariance, i.e., we fix $\vSigma^*_i=\vSigma^*$ for all $i$. First, we assume that the mixing weights $w_i$ are also fixed to known values. Under this model, which we call {\em Model 1}, EM finds a stationary point of the log-likelihood function in the parameter space of component means $(\vtheta_1,\ldots, \vtheta_k)$. Next, we over-parameterize Model 1 as follows. Despite the fact that the weights fixed in Model 1, we now pretend that they are not fixed. This gives a second model, which we call {\em Model 2}. Parameter estimation for Model 2 requires EM to estimate the mixing weights in addition to the component means. Finding the global maximizer of the log-likelihood over this enlarged parameter space is seemingly more difficult for Model 2 than it is for Model 1, and perhaps needlessly so. However, in this paper we present theoretical and empirical evidence to the contrary.
\begin{enumerate}

  \item For mixtures of two symmetric Gaussians (i.e., $k=2$ and $\vtheta^*_1=-\vtheta^*_2$), we prove that EM for Model 2 converges to the global maximizer of the log-likelihood objective with almost any initialization of the mean parameters, while EM for Model 1 will fail to do so for many choices of $(w^*_1,w^*_2)$. These results are established for idealized executions of EM in an infinite sample size limit, which we complement with finite sample results.

  \item
    We prove that the spurious local maxima in the (population) log-likelihood objective for Model 1 are eliminated in the objective for Model 2.

  \item We present an empirical study to show that for more general mixtures of Gaussians, with a variety of model parameters and sample sizes, EM for Model 2 has higher probability to find the MLE than Model 1 under random initializations.
\end{enumerate}

\paragraph{Related work.}
Since Dempster's 1977 paper \citep{dempster1977ml}, the EM algorithm has become one of the most popular algorithms to find the MLE for mixture models. Due to its popularity, the convergence analysis of EM has attracted researchers' attention for years. Local convergence of EM has been shown by \citet{wu1983convergence, xu1996convergence, tseng2004analysis,chretien2008em}. Further, for certain models and under various assumptions about the initialization, EM has been shown to converge to the MLE~\citep{redner1984mixture,balakrishnan2014statistical, yale2016statistical, yan2017convergence}. Typically, the initialization is required to be sufficiently close to the true parameter values of the data-generating distribution. Much less is known about global convergence of EM, as the landscape of the log-likelihood function has not been well-studied. For GMMs, \citet{xu2016global} and \citep{daskalakis2017ten} study mixtures of two Gaussians with equal weights and show that the log-likelihood objective has only two global maxima and one saddle point; and if EM is randomly initialized in a natural way, the probability that EM converges to this saddle point is zero. (Our Theorem~\ref{thm:main} generalizes these results.)
It is known that for mixtures of three or more Gaussians, global convergence is not generally possible~\citep{jin2016local}.

The value of over-parameterization for local or greedy search algorithms that aim to find a global minimizer of non-convex objectives has been rigorously established in other domains. Matrix completion is a concrete example: the goal is to recover of a rank $r \ll n$ matrix $M\in \bbR^{n\times n}$ from observations of randomly chosen entries~\citep{candes2009exact}. A direct approach to this problem is to find the matrix $X\in \bbR^{n\times n}$ of minimum rank that is consistent with the observed entries of $M$. However, this optimization problem is NP-hard in general, despite the fact that there are only $2nr-r^2 \ll n^2$ degrees-of-freedom. An indirect approach to this matrix completion problem is to find a matrix $X$ of smallest nuclear norm, subject to the same constraints; this is a convex relaxation of the rank minimization problem. By considering all $n^2$ degrees-of-freedom, \citet{candes2010power} show that the matrix $M$ is exactly recovered via nuclear norm minimization as soon as $\Omega(nr\log^6 n)$ entries are observed (with high probability). Notably, this combination of over-parameterization with convex relaxation works well in many other research problems such as sparse-PCA \citep{d2005direct} and compressive sensing~\citep{donoho2006compressed}. However, many problems (like ours) do not have a straightforward convex relaxation. Therefore, it is important to understand how over-parameterization can help one solve a non-convex problem other than convex relaxation. 

Another line of work in which the value of over-parameterization is observed is in deep learning. It is conjectured that the use of over-parameterization is the main reason for the success of local search algorithms in learning good  parameters for neural nets~\citep{livni2014computational, safran2017spurious}. Recently, \citet{haeffele2015global,nguyen2017lossa,nguyen2017lossb, soltani2018towards,du2018power} confirm this observation for many neural networks such as feedforward and convolutional neural networks.

\section{Theoretical results}

In this section, we present our main theoretical results concerning EM and two-component Gaussian mixture models.

\subsection{Sample-based EM and Population EM}\label{sec:N&O}

Without loss of generality, we assume $\vSigma^*=\vI$. We consider the following Gaussian mixture model:
\begin{eqnarray}
\vy_1,\dotsc,\vy_n&\stackrel{\text{i.i.d.}}{\sim}& w^*_1 \mathcal{N}(\vtheta^*, \vI)+w^*_2 \mathcal{N}(-\vtheta^*, \vI).
  \label{eq:modeltheory}
\end{eqnarray}
The mixing weights $w^*_1$ and $w^*_2$ are \emph{fixed} (i.e., assumed to be known). Without loss of generality, we also assume that $w^*_1 \geq w^*_2>0$ (and, of course, $w^*_1+w^*_2=1$). The only parameter to estimate is the mean vector $\vtheta^*$.
The EM algorithm for this model uses the following iterations:
\begin{eqnarray}
\hvpt{t+1}&=&\frac{1}{n} \sum_{i=1}^n \left[\frac{w^*_1e^{\dotp{\vy_i,\hvpt{t}}}-w^*_2e^{-\dotp{\vy_i,\hvpt{t}}}}{w^*_1e^{\dotp{\vy_i,\hvpt{t}}}+w^*_2e^{-\dotp{\vy_i,\hvpt{t}}}}\vy_i\right].\label{eq:update1thetas}
\end{eqnarray}    
We refer to this algorithm as \SBEM{1}: it is the EM algorithm one would normally use when the mixing weights are known. In spite of this, we also consider an EM algorithm that pretends that the weights are not known, and estimates them alongside the mean parameters. We refer to this algorithm as \SBEM{2}, which uses the following iterations:
\begin{eqnarray}
\hpw{t+1}_1&=& \frac{1}{n} \sum_{i=1}^n\left[\frac{\hpw{t}_1e^{\dotp{\vy_i,\hvpt{t}}}}{\hpw{t}_1e^{\dotp{\vy_i,\hvpt{t}}}+\hpw{t}_2e^{-\dotp{\vy_i,\hvpt{t}}}}\right]\=1-\hpw{t+1}_2. \n\\
\hvpt{t+1}&=&\frac{1}{n} \sum_{i=1}^n \left[\frac{\hpw{t}_1e^{\dotp{\vy_i,\hvpt{t}}}-\hpw{t}_2e^{-\dotp{\vy_i,\hvpt{t}}}}{\hpw{t}_1e^{\dotp{\vy_i,\hvpt{t}}}+\hpw{t}_2e^{-\dotp{\vy_i,\hvpt{t}}}}\vy_i\right].
\label{eq:update2thetas}
\end{eqnarray}
This is the EM algorithm for a different Gaussian mixture model in which the weights $w^*_1$ and $w^*_2$ are not fixed (i.e., unknown), and hence must be estimated. Our goal is to study the global convergence properties of the above two EM algorithms on data from the \emph{first} model, where the mixing weights are, in fact, known.

We study idealized executions of the EM algorithms in the large sample limit, where the algorithms are modified to be computed over an infinitely large i.i.d.~sample drawn from the mixture distribution in~\eqref{eq:modeltheory}. Specifically, we replace the empirical averages in \eqref{eq:update1thetas} and \eqref{eq:update2thetas} with the expectations with respect to the mixture distribution. We obtain the following two modified EM algorithms, which we refer to as \PEM{1} and \PEM{2}:
\begin{itemize}
  \item \PEM{1}:

\begin{eqnarray}
\vpt{t+1}&=&\bbE_{\vy\sim f^*}\left[\frac{w^*_1e^{\dotp{\vy,\vpt{t}}}-w^*_2e^{-\dotp{\vy,\vpt{t}}}}{w^*_1e^{\dotp{\vy,\vpt{t}}}+w^*_2e^{-\dotp{\vy,\vpt{t}}}}\vy\right]\ =: \  H(\vpt{t};\vtheta^*,w^*_1),\label{eq:H}
\end{eqnarray}    
where $f^*=f^*(\vtheta^*,w^*_1)$ here denotes the true distribution of $\vy_i$ given in \eqref{eq:modeltheory}. 

\item \PEM{2}: Set $\pw{0}_1 = \pw{0}_2= 0.5$\footnote{Using equal initial weights is a natural way to initialize EM when the weights are unknown.}, and run

\begin{eqnarray}
\pw{t+1}_1&=& \bbE_{\vy\sim f^*}\left[\frac{\pw{t}_1e^{\dotp{\vy,\vpt{t}}}}{\pw{t}_1e^{\dotp{\vy,\vpt{t}}}+\pw{t}_2e^{-\dotp{\vy,\vpt{t}}}}\right]\ =: \ G_{w}(\vpt{t},\pw{t};\vtheta^*,w^*_1)
  \label{eq:update2w}
  \\
&=&1-\pw{t+1}_2 .
\nonumber
\end{eqnarray}    

\begin{eqnarray}
\vpt{t+1}&=&\bbE_{\vy\sim f^*}\left[\frac{\pw{t}_1e^{\dotp{\vy,\vpt{t}}}-\pw{t}_2e^{-\dotp{\vy,\vpt{t}}}}{\pw{t}_1e^{\dotp{\vy,\vpt{t}}}+\pw{t}_2e^{-\dotp{\vy,\vpt{t}}}}\vy\right]\ =: \ G_{\theta}(\vpt{t},\pw{t};\vtheta^*,w^*_1).
\label{eq:update2theta}
\end{eqnarray}    
\end{itemize}
As $n \rightarrow \infty$, we can show the performance of \SBEM{\star} converges to that of the \PEM{\star} in probability. This argument has been used rigorously in many previous works on EM~\citep{balakrishnan2014statistical,xu2016global,yale2016statistical,daskalakis2017ten}. The main goal of this section, however, is to study the dynamics of \PEM{1} and \PEM{2}, and the landscape of the log-likelihood objectives of the two models. 

\subsection{Main theoretical results}\label{sec:theory}

Let us first consider the special case $w^*_1= w^*_2= 0.5$. Then, it is straightforward to show that $\pw{t}_1=\pw{t}_2=0.5$ for all $t$ in \PEM{2}. Hence, \PEM{2} is equivalent to \PEM{1}. Global convergence of $\vpt{t}$ to $\vtheta^*$ for this case was recently established by~\citet[Theorem~1]{xu2016global} for almost all initial $\vpt{0}$ (see also~\citep{daskalakis2017ten}).

We first show that the same global convergence may \emph{not} hold for \PEM{1} when $w^*_1 \neq w^*_2$.

\begin{figure}
\begin{center}
\includegraphics[width=0.75\textwidth]{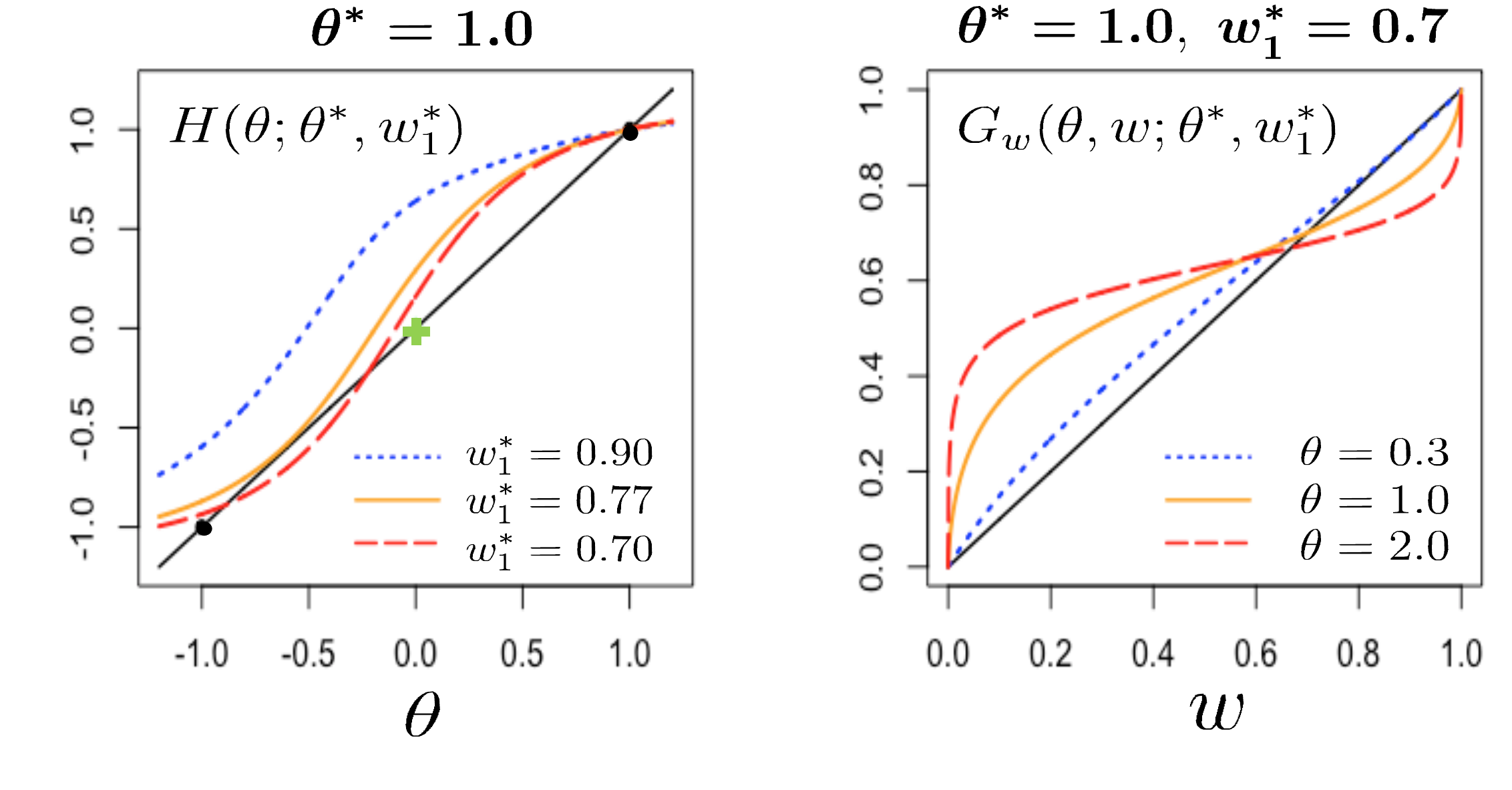}
  \vspace{-0.8cm}
\end{center}
\caption{Left panel: we show the shape of iterative function $H(\theta; \theta^*, w^*_1)$ with $\theta^*=1$ and different values of $w^*_1\in \{0.9, 0.77,0.7\}$. The green plus $+$ indicates the origin $(0,0)$ and the black points indicate the correct values $(\theta^*,\theta^*)$ and $(-\theta^*,-\theta^*)$. We observe that as $w^*_1$ increases, the number of fixed points goes down from $3$ to $2$ and finally to $1$. Further, when there exists more than one fixed point, there is one stable incorrect fixed point in $(-\theta^*,0)$. Right panel: we show the shape of iterative function $G_{w}(\theta,w_1; \theta^*, w^*_1)$ with $\theta^*=1, w^*_1=0.7$ and different values of $\theta \in \{0.3,1,2\}$. We observe that as $\theta$ increases, $G_w$ becomes from a concave function to a concave-convex function. Further, there are at most three fixed points and there is only one stable fixed point.}
\label{fig:Model1H}
\end{figure}

\begin{theorem}\label{thm:Model1}
  Consider \PEM{1} in dimension one (i.e., $\theta^*\in \bbR$). For any $\theta^*>0$, there exists $\delta>0$, such that given $w_1^*\in (0.5,0.5+\delta)$ and initialization $\pt{0}\leq -\theta^*$, the \PEM{1} estimate $\pt{t}$ converges to a fixed point $\theta_{\rm wrong}$ inside $(-\theta^*,0)$.
\end{theorem}
This theorem, which is proved in Appendix~\ref{sec:proofthm2}, implies that if we use random initialization, \PEM{1} may converge
to the wrong fixed point with constant probability. We illustrate this in Figure~\ref{fig:Model1H}.
The iterates of \PEM{1}
converge to a fixed point of the function $\theta \mapsto H(\theta; \theta^*, w^*_1)$ defined in \eqref{eq:H}. We have plotted this function for several different values of $w^*_1$ in the left panel of Figure~\ref{fig:Model1H}. When $w^*_1$ is close to $1$, $H(\theta; \theta^*, w^*_1)$ has only one fixed point and that is at $\theta= \theta^*$. Hence, in this case, the estimates produced by \PEM{1} converge to the true $\theta^*$. However, when we decrease the value of $w^*_1$ below a certain threshold (which is numerically found to be approximately $0.77$ for $\theta^*=1$), two other fixed points of $H(\theta; \theta^*, w^*_1)$ emerge. These new fixed points are foils for \PEM{1}.

From the failure of \PEM{1}, one may expect the over-parameterized \PEM{2} to fail as well. Yet, surprisingly, our second theorem proves the opposite is true: \PEM{2} has global convergence even when $w^*_1 \neq w^*_2$.

\begin{theorem}\label{thm:main}
  For any $w_1^*\in [0.5,1)$, the \PEM{2} estimate $(\vtheta^t,\pw{t})$ converges to either $(\vtheta^*,w_1^*)$ or $(-\vtheta^*,w_2^*)$ with any initialization $\vpt{0}$ except on the hyperplane $\dotp{\vpt{0},\vtheta^*}=0$. Furthermore, the convergence speed is geometric after some finite number of iterations, i.e., there exists a finite number $T$ and constant $\rho\in (0,1)$ such that the following hold.
  \begin{itemize}
\item If $\dotp{\vpt{0},\vtheta^*}>0$, then for all $t>T$,
\begin{align*}
	\|\vpt{t+1}-\vtheta^*\|^2+|\pw{t+1}_1-w_1^*|^2
	&\leq
	\rho^{t-T}\left(\|\vpt{T}-\vtheta^*\|^2+(\pw{T}_1-w_1^*)^2\right)
	.
\end{align*}
\item If $\dotp{\vpt{0},\vtheta^*}<0$, then for all $t>T$,
\begin{align*}
	\|\vpt{t+1}+\vtheta^*\|^2+|\pw{t+1}_1-w_2^*|^2
	&\leq
	\rho^{t-T}\left(\|\vpt{T}+\vtheta^*\|^2+(\pw{T}_1-w_2^*)^2\right)
	.
\end{align*}
\end{itemize}

\end{theorem}

Theorem~\ref{thm:main} implies that if we use random initialization for $\vpt{0}$, with probability one, the \PEM{2} estimates converge to the true parameters.


The failure of \PEM{1} and success of \PEM{2} can be explained intuitively. Let $C_1$ and $C_2$, respectively, denote the true mixture components with parameters $(w_1^*,\theta^*)$ and $(w_2^*,-\theta^*)$. Due to the symmetry in \PEM{1}, we are assured that among the two estimated mixture components, one will have a positive mean, and the other will have a negative mean: call these $\hat C_+$ and $\hat C_-$, respectively.
Assume $\theta^*>0$ and $w_1^*>0.5$, and consider initializing the \PEM{1} with $\theta^{\langle 0 \rangle} := -\theta^*$. This initialization incorrectly associates $\hat C_-$ with the larger weight $w_1^*$ instead of the smaller weight $w_2^*$. This causes, in the E-step of EM,
the component $\hat C_-$ to become ``responsible'' for an overly large share of the overall probability mass, and in particular an overly large share of the mass from $C_1$ (which has a positive mean). Thus, in the M-step of EM, when the mean of the estimated component $\hat C_-$ is updated, it is pulled rightward towards $+\infty$. It is possible that this rightward pull would cause the estimated mean of $\hat C_-$ to become positive---in which case the roles of $\hat C_+$ and $\hat C_-$ would switch---but this will not happen as long as $w_1^*$ is sufficiently bounded away from $1$ (but still $>0.5$).\footnote{When $w_1^*$ is indeed very close to $1$, then almost all of the probability mass of the true distribution comes from $C_1$, which has positive mean. So, in the M-step discussed above, the rightward pull of the mean of $\hat C_-$ may be so strong that the updated mean estimate becomes positive. Since the model enforces that the mean estimates of $\hat C_+$ and $\hat C_-$ be negations of each other, the roles of $\hat C_+$ and $\hat C_-$ switch, and now it is $\hat C_+$ that becomes associated with the larger mixing weight $w_1^*$. In this case, owing to the symmetry assumption, \PEM{1} may be able to successfully converge to $\theta^*$. We revisit this issue in the numerical study, where the symmetry assumption is removed.} The result is a bias in the estimation of $\theta^*$, thus explaining why the \PEM{1} estimate converges to some $\theta_{\text{wrong}} \in (-\theta^*,0)$ when $w_1^*$ is not too large.


%

Our discussion confirms that one way \PEM{1} may fail (in dimension one) is if it is initialized with $\theta^{\langle 0 \rangle}$ having the ``incorrect'' sign (e.g., $\theta^{\langle 0 \rangle} = -\theta^*$). On the other hand, the performance of \PEM{2} does not depend on the sign of the initial $\theta^{\langle 0 \rangle}$. Recall that the estimates of \PEM{2} converge to the fixed points of the mapping $\mathcal{M}: (\vtheta, w_1) \mapsto (G_{\theta}(\vtheta,w_1;\vtheta^*,w^*_1), G_{w}(\vtheta,w_1;\vtheta^*,w^*_1) )$, as defined in~\eqref{eq:update2w} and~\eqref{eq:update2theta}. One can check that for all $\vtheta,w_1,\vtheta^*,w^*_1$, we have
\begin{equation}
  \begin{aligned}
    G_{\theta}(\vtheta,w_1;\vtheta^*,w^*_1)+G_{\theta}(-\vtheta,w_2;\vtheta^*,w^*_1)&=0,\\
    G_{w}(\vtheta,w_1;\vtheta^*,w^*_1)+G_{w}(-\vtheta,w_2;\vtheta^*,w^*_1)&=1.
  \end{aligned}
  \label{eq:sign}
\end{equation}
Hence, $(\vtheta,w_1)$ is a fixed point of $\mathcal{M}$ if and only if $(-\vtheta,w_2)$ is a fixed point of $\mathcal{M}$ as well. Therefore, \PEM{2} is insensitive to the sign of the initial $\theta^{\langle 0 \rangle}$. This property can be extended to mixtures of $k>2$ Gaussians as well. In these cases, the performance of EM for Model 2 is insensitive to permutations of the component parameters. Hence, because of this nice property, as we will confirm in our simulations, when the mixture components are well-separated, EM for Model 2 performs well for most of the initializations, while EM for Model 1 fails in many cases. 

  One limitation of our permutation-free explanation is that the argument only holds when the weights in \PEM{2} are initialized to be uniform. However, the benefits of over-parameterization are not limited to this case. Indeed, when we compare the landscapes of the log-likelihood objective for (the mixture models corresponding to) \PEM{1} and \PEM{2}, we find that over-parameterization eliminates spurious local maxima that were obstacles for \PEM{1}.
%
\begin{theorem}\label{thm:landscape}
 For all $w_1^*\neq 0.5$, the log-likelihood objective optimized by \PEM{2} has only one saddle point $(\vtheta,w_1) = (\v0,1/2)$ and no local maximizers
besides the two global maximizers $(\vtheta,w_1) = (\vtheta^*,w_1^*)$ and $(\vtheta,w_1) = (-\vtheta^*,w_2^*)$.
\end{theorem}
The proof of this theorem is presented in Appendix~\ref{sec:landscape}. 


\begin{remark}
Consider the landscape of the log-likelihood objective for \PEM{2} and the point $(\theta_{\rm wrong},w_1^*)$, where $\theta_{\rm wrong}$ is the local maximizer suggested by Theorem~\ref{thm:Model1}. Theorem~\ref{thm:landscape} implies that we can still easily escape this point due to the non-zero gradient in the direction of $w_1$ and thus $(\theta_{\rm wrong},w_1^*)$ is not even a saddle point. We emphasize that this is exactly the mechanism that we have hoped for the purpose and benefit of over-parameterization. 
\end{remark}
\begin{remark}
  Note that although $(\vtheta,w_1) = ((w_1^*-w_2^*)\vtheta^*,1)$ or $((w_2^*-w_1^*)\vtheta^*,0)$ are the two fixed points for \PEM{2} as well, they are not the first order stationary points of the log-likelihood objective if $w_1^*\neq 0.5$. 
\end{remark}

Finally, to complete the analysis of EM for the mixtures of two Gaussians, we present the following result that applies to \SBEM{2}.
\begin{theorem}\label{thm:fsa}
  Let $(\hvpt{t},\hpw{t}_1)$ be the estimates of \SBEM{2}. Suppose $\hvpt{0}=\vpt{0}, \hpw{0}_1=\pw{0}_1=\frac{1}{2}$ and $\dotp{\vpt{0},\vtheta^*}\neq 0$. Then we have
\begin{equation}
	\limsup_{t\rightarrow \infty} \|\hvpt{t}-\vpt{t}\|
	\ \rightarrow \
	0
	\kand
	\limsup_{t\rightarrow\infty} |\hpw{t}_1-\pw{t}_1|
	\ \rightarrow \
	0
	\quad \text{as $n\to\infty$}
	\,,	\n
\end{equation}
where convergence is in probability.
\end{theorem} 
The proof of this theorem uses the same approach as~\citet{xu2016global} and is presented in Appendix~\ref{sec:fsa}.

\subsection{Roadmap of the proof for Theorem \ref{thm:main}}\label{sec:mainproof}

Our first lemma, proved in Appendix~\ref{sec:proofinitialarea}, confirms that if $\dotp{\vpt{0},\vtheta^*}>0$, then $\dotp{\vpt{t},\vtheta^*}>0$ for every $t$ and $\pw{t}_1\in (0.5,1)$. In other words, the estimates of the \PEM{2} remain in the correct hyperplane, and the weight moves in the right direction, too. 

\begin{lemma} \label{lem:initialarea} 
If $\dotp{\vpt{0},\vtheta^*}>0$, we have $\dotp{\vpt{t},\vtheta^*}>0, \pw{t}_1\in (0.5,1)$ for all $t\geq 1$. Otherwise, if $\dotp{\vpt{0},\vtheta^*}<0$, we have $\dotp{\vpt{t},\vtheta^*}<0, \pw{t}_1\in (0,0.5)$ for all $t\geq 1$.
\end{lemma}
On account of Lemma~\ref{lem:initialarea} and the invariance
in~\eqref{eq:sign}, we can assume without loss of generality that
$\dotp{\vpt{t},\vtheta^*}>0$ and $\pw{t}_1\in (0.5,1)$ for all $t\geq 1$.

Let $d$ be the dimension of $\vtheta^*$. We reduce the $d>1$ case to the $d=1$ case. This achieved by proving that the angle between the two vectors $\vpt{t}$ and $\vtheta^*$ is a decreasing function of $t$ and converges to $0$. The details appear in Appendix~\ref{sec:reduceto1}. Hence, in the rest of this section we focus on the proof of Theorem \ref{thm:main} for $d=1$.

\begin{figure}
\begin{center}
\includegraphics[width=0.65\textwidth]{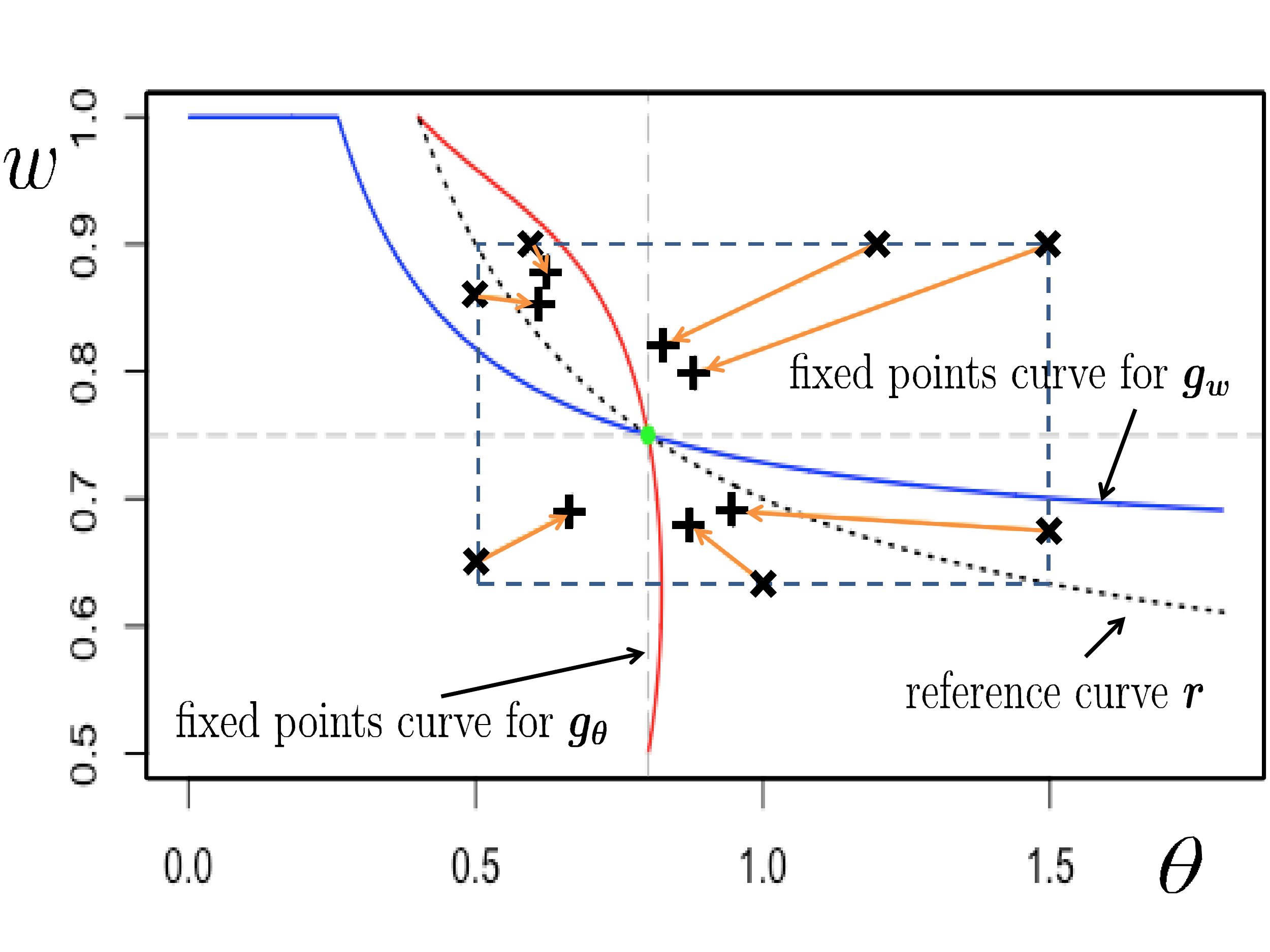}
  \vspace{-0.8cm}
\end{center}
\caption{The functions $g_{\theta}$ and $g_w$ are shown with red and blue lines respectively. The green point at the intersections of the three curves is the correct convergence point $(\theta_{\star},w_{\star})$. The black dotted curve shows the reference curve $r$. The cross points $\times$ are the possible initializations and the plus points $+$ are the corresponding positions after the first iteration. By the geometric relations between the three curves, the iterations have to converge to $(\theta_{\star},w_{\star})$}
\label{fig:convergef}
\end{figure}

Let $g_{\theta}(\theta,w_1)$ and $g_w(\theta,w_1)$ be the shorthand for the two update functions $G_{\theta}$ and $G_w$ defined in \eqref{eq:update2w} and \eqref{eq:update2theta} for a fixed $(\theta^*,w^*_1)$. To prove that $\{(\pt{t},\pw{t})\}$ converges to the fixed point $(\theta_{\star},w_{\star})$, we establish the following claims:
\begin{itemize}
  \item[C.1] There exists a set $S=(a_{\theta},b_{\theta})\times (a_w,b_w)\in
    \bbR^2$, where $a_{\theta},b_{\theta}\in \bbR\cup \{\pm\infty\}$ and
    $a_w,b_w\in \bbR$, such that $S$ contains point
    $(\theta_{\star},w_{\star})$ and point
    $(g_{\theta}(\theta,w_1),g_w(\theta,w_1))\in S$ for all $(\theta,w_1)\in
    S$. Further, $g_{\theta}(\theta,w_1)$ is a non-decreasing function of
    $\theta$ for a given $w_1\in(a_w,b_w)$ and $g_{w}(\theta,w_1)$ is a
    non-decreasing function of $w$ for a given
    $\theta\in(a_{\theta},b_{\theta})$,

  \item[C.2] There is a \emph{reference curve} $r \colon [a_w,b_w] \rightarrow
    [a_{\theta},b_{\theta}]$ defined on $\bar{S}$ (the closure of $S$) such
    that:

  \begin{itemize}
    \item[C.2a] $r$ is continuous, decreasing, and passes through point $(\theta_{\star},w_{\star})$, i.e., $r(w_{\star})=\theta_{\star}.$ 
    \item[C.2b] 
      Given $\theta \in (a_{\theta},b_{\theta})$, function $w \mapsto g_w(\theta,w)$ has a stable fixed point in $[a_w,b_w]$. Further, any stable fixed point $w_s$ in $[a_w,b_w]$ or fixed point $w_s$ in $(a_w,b_w)$ satisfies the following:
      \begin{itemize}
        \item If $\theta<\theta_{\star}$ and $\theta\geq r(b_w)$, then $r^{-1}(\theta)>w_s>w_{\star}$. 
        \item If $\theta=\theta_{\star}$, then $r^{-1}(\theta)=w_s=w_{\star}$.  
        \item If $\theta>\theta_{\star}$ and $\theta\leq r(a_w)$, then $r^{-1}(\theta)<w_s<w_{\star}$.  
      \end{itemize}  
      \item[C.2c] Given $w\in [a_w,b_w]$, function $\theta \mapsto g_{\theta}(\theta,w)$ has a stable fixed point in $[a_{\theta},b_{\theta}]$. Further, any stable fixed point $\theta_s$ in $[a_{\theta},b_{\theta}]$ or fixed point $\theta_s$ in $(a_{\theta},b_{\theta})$ satisfies the following:
      \begin{itemize}
        \item If $w_1<w_{\star}$, then $r(w)> \theta_s>\theta_{\star}$.  
        \item If $w_1=w_{\star}$, then $r(w)=\theta_s=\theta_{\star}$.  
        \item If $w_1>w_{\star}$, then $r(w)< \theta_s<\theta_{\star}$.  
      \end{itemize} 
  \end{itemize}
\end{itemize}
We explain C.1 and C.2 in Figure \ref{fig:convergef}. Heuristically, we expect $(\theta^*,w^*_1)$ to be the only fixed point of the mapping $(\theta, w)\mapsto (g_{\theta}(\theta,w), g_{w}(\theta,w))$, and that $(\pt{t},\pw{t})$ move toward this fixed point. Hence, we can prove the convergence of the iterates by showing certain geometric relationships between the curves of fixed points of the two functions. Hence, C.1 helps us to bound the iterates on the area that such nice geometric relations exist, and the reference curve $r$ and C.2 are the tools to help us mathematically characterizing the geometric relations shown in the figure. Indeed, the next lemma implies that $C.1$ and $C.2$ are sufficient to show the convergence to the right point $(\theta_{\star},w_{\star})$:
\begin{lemma}[Proved in Appendix~\ref{sec:proofconverge}]\label{lem:converge}
Suppose continuous functions $g_{\theta}(\theta,w),g_w(\theta,w)$ satisfy $C.1$ and $C.2$,  then there exists a continuous mapping $m:\bar{S}\rightarrow [0,\infty)$ such that $(\theta_{\star},w_{\star})$ is the only solution for $m(\theta,w)=0$ on $\bar{S}$, the closure of $S$
. Further, if we initialize $(\pt{0},\pw{0})$ in $S$, the sequence $\{(\pt{t},\pw{t})\}_{t\geq 0}$ defined by
\begin{eqnarray}
\pt{t+1}&=&g_{\theta}(\pt{t},\pw{t}), \kand  \pw{t+1}\= g_w(\pt{t},\pw{t}),\n
\end{eqnarray}  
satisfies that $m(\pt{t},\pw{t})\downarrow 0$, and therefore $(\pt{t},\pw{t})$ converges to $(\theta_{\star},w_{\star})$.
\end{lemma}  

In our problem, we set $a_w=0.5, b_w=1, a_{\theta}=0,b_{\theta}=\infty$ and $(\theta_{\star},w_{\star})=(\theta^*,w^*_1)$. Then according to Lemma \ref{lem:initialarea} and monotonic property of $g_{\theta}$ and $g_w$, C.1 is satisfied.

To show C.2, we first define the reference curve $r$ by
\begin{eqnarray}
r(w_1): = \frac{w^*_1-w^*_2}{w_1-w_2}\theta^* \= \frac{2w^*_1-1}{2w_1-1}\theta^*, \quad \quad \forall w_1\in (0.5,1],w_2=1-w_1.\label{eq:r}
\end{eqnarray}
The claim C.2a holds by construction.
To show C.2b, we establish an even stronger property of the weights update function $g_w(\theta,w)$: for any fixed $\theta>0$, the function $w_1 \mapsto g_w(\theta, w_1)$ has at most one other fixed point besides $w_1=0$ and $w_1=1$, and most importantly, it has only one unique stable fixed point. This is formalized in the following lemma.
\begin{lemma}[Proved in Appendix~\ref{sec:proofshapeg}] \label{lem:shapeg}
For all $\theta>0$, there are at most three fixed points for $g_w(\theta,w_1)$ with respect to $w_1$. Further, there exists an unique stable fixed point $F_w(\theta)\in (0,1]$, i.e., (i) $F_w(\theta)= g_w(\theta,F_w(\theta))$ and (ii) for all $w_1\in (0,1)$, we have
\begin{equation}
  g_w(\theta,w_1) < w_1 \Leftrightarrow w_1 < F_w(\theta)
  \quad \text{and} \quad
  g_w(\theta,w_1) > w_1 \Leftrightarrow w_1 > F_w(\theta)
  .
  \label{eq:shapeofg}
\end{equation}
\end{lemma}
We explain Lemma \ref{lem:shapeg} in Figure \ref{fig:Model1H}. Note that, in the figure, we observe that $g_w$ is an increasing function with $g_w(\theta,0)=0$ and $g_w(\theta,1)=1$. Further, it is either a concave function, it is piecewise concave-then-convex\footnote{There exists $\tilde{w}\in (0,1)$ such that $g_w(\theta,w)$ is concave in $[0,\tilde{w}]$ and convex in $[\tilde{w},1]$.}. Hence, we know if $\partial g_{w}(\theta,w_1)/\partial w_1|_{w_1=1}$ is at most $1$, the only stable fixed point is $w_1=1$, else if the derivative is larger than 1, there exists only one fixed point in (0,1) and it is the only stable fixed point. The complete proof for C.2b is shown in Appendix~\ref{sec:proofc2bfirst}.

The final step to apply Lemma \ref{lem:converge} is to prove C2.c. However, $(\theta,w_1)=((2w^*_1-1)\theta^*,1)$ is a point on the reference curve $r$ and $\theta=(2w^*_1-1)\theta^*$ is a stable fixed point for $g_{\theta}(\theta,1)$. This violates C.2c. To address this issue, since we can characterize the shape and the number of fixed points for $g_w$, by typical uniform continuity arguments, we can find $\delta,\epsilon>0$ such that the adjusted reference curve $r_{adj}(w):=r(w)-\epsilon\cdot \max(0,w-1+\delta)$ satisfies C.2a and C.2b. Then we can prove that the adjusted reference curve $r_{adj}(w)$ satisfies C2.c; see Appendix~\ref{sec:proofc2c}.

\section{Numerical results}
\label{sec:simulation}

In this section, we present numerical results that show the value of over-parameterization in some mixture models not covered by our theoretical results.

\subsection{Setup}

Our goal is to analyze the effect of the sample size, mixing weights, and the number of mixture components on the success of the two EM algorithms described in Section~\ref{sec:N&O}.

We implement EM for both Model 1 (where the weights are assumed to be known) and Model 2 (where the weights are not known), and run the algorithm multiple times with random initial mean estimates. We compare the two versions of EM by their (empirical) success probabilities, which we denote by $P_1$ and $P_2$, respectively. Success is defined in two ways, depending on whether EM is run with a finite sample, or with an infinite-size sample (i.e., the population analogue of EM).

When EM is run using a finite sample, we do not expect recover the $\vtheta^*_i \in \bbR^d$ exactly.
Hence, success is declared when the $\vtheta^*_i$ are recovered up to some expected error, according to the following measure:
\begin{eqnarray}
\text{error}\ =\  \min_{\pi\in \Pi} \lim_{t\rightarrow \infty}\sum_{i=1}^kw^*_i\|\vpt{t}_{\pi(i)}-\vtheta^*_i\|^2,\label{eq:err}
\end{eqnarray}
where $\Pi$ is the set of all possible permutations on $\{1,\ldots,k\}$. We declare success if the error is at most $C_{\epsilon}/n$, where $C_\epsilon := 4\cdot \text{Tr}(\vW^*\mathcal{I}^{-1}(\vTheta^*))$. Here, $\vW^*$ is the diagonal matrix whose diagonal is $(w^*_1,\ldots,w^*_1,\dotsc,w^*_k,\ldots,w^*_k)\in \bbR^{kd}$, where each $w^*_i$ is repeated $d$ times, and $\mathcal{I}(\vTheta^*)$ is the Fisher Information at the true value $\vTheta^* := (\vtheta^*_1,\dotsc,\vtheta^*_k)$. We adopt this criteria since it is well known that the MLE asymptotically converges to $\mathcal{N}(\theta^*,\mathcal{I}^{-1}(\vTheta^*)/n)$. Thus, constant $4\approx 1.96^2$ indicates an approximately $95\%$ coverage.

When EM is run using an infinite-size sample, we declare EM successful when the error defined in~\eqref{eq:err} is at most $10^{-7}$.

\subsection{Mixtures of two Gaussians}\label{sec:twoG}

We first consider mixtures of two Gaussians in one dimension, i.e., $\theta^*_1,\theta^*_2 \in \bbR$. Unlike in our theoretical analysis, the mixture components are not constrained to be symmetric about the origin. For simplicity, we always let $\theta^*_1=0$, but this information is not used by EM. Further, we consider sample size $n\in \{1000,\infty\}$, separation $\theta^*_2=|\theta^*_2-\theta^*_1|\in\{1,2,4\}$, and mixing weight $w^*_1\in\{0.52, 0.7,0.9\}$; this gives a total of 18 cases. For each case, we run EM with $2500$ random initializations and compute the empirical probability of success. When $n=1000$, the initial mean parameter is chosen uniformly at random from the sample. When $n=\infty$, the initial mean parameter is chosen uniformly at random from the rectangle $[-2,\theta^*_2+2]\times [-2,\theta^*_2+2]$.


A subset of the success probabilities are shown in Table~\ref{tab:1}; see Appendix~\ref{sec:tables} for the full set of results. Our simulations lead to the following empirical findings about the behavior of EM on data from well-separated mixtures ($|\theta^*_1-\theta^*_2|\geq1$). First, for $n=\infty$, EM for Model 2 finds the MLE almost always ($P_2=1$), while EM for Model 1 only succeeds about half the time ($P_1\approx 0.5$). Second, for smaller $n$, EM for Model 2 still has a higher chance of success than EM for Model 1, except when the weights $w^*_1$ and $w^*_2$ are almost equal. When $w^*_1\approx w^*_2 \approx 1/2$, the bias in Model 1 is not big enough to stand out from the error due to the finite sample, and hence Model 1 is more preferable. Notably, unlike the special model in \eqref{eq:modeltheory}, highly unbalanced weights do not help EM for Model 1 due to the lack of the symmetry of the component means (i.e., we may have $\theta^*_1 + \theta^*_2 \neq 0$).

We conclude that over-parameterization helps EM if the two mixture components are well-separated and the mixing weights are not too close.
%

\subsection{Mixtures of three or four Gaussians}\label{sec:3G} 

We now consider a setup with mixtures of three or four Gaussians.
Specifically, we consider the following four cases, each using a larger sample size of $n=2000$:
\begin{itemize}
\item Case 1, mixture of three Gaussians on a line: $\vtheta^*_1=(-3,0)$, $\vtheta^*_2=(0,0)$, $\vtheta^*_3=(2,0)$ with weights $w^*_1=0.5,w^*_2=0.3,w^*_3=0.2$. 
\item Case 2, mixture of three Gaussians on a triangle: $\vtheta^*_1=(-3,0)$, $\vtheta^*_2=(0,2)$, $\vtheta^*_3=(2,0)$ with weights $w^*_1=0.5,w^*_2=0.3,w^*_3=0.2$.
\item Case 3, mixture of four Gaussians on a line: $\vtheta^*_1=(-3,0)$, $\vtheta^*_2=(0,0)$, $\vtheta^*_3=(2,0)$, $\vtheta^*_4=(5,0)$ with weights $w^*_1=0.35,w^*_2=0.3,w^*_3=0.2,w^*_4=0.15$.
\item Case 4, mixture of four Gaussians on a trapezoid:  $\vtheta^*_1=(-3,0)$, $\vtheta^*_2=(-1,2)$, $\vtheta^*_3=(2,0)$, $\vtheta^*_4=(2,2)$ with weights $w^*_1=0.35,w^*_2=0.3,w^*_3=0.2,w^*_4=0.15$.
\end{itemize}
The other aspects of the simulations are the same as in the previous subsection.

The results are presented in Table \ref{tab:1}. From the table, we confirm that EM for Model 2 (with unknown weights) has a higher success probability than EM for Model 1 (with known weights). Therefore, over-parameterization helps in all four cases.

\subsection{Explaining the disparity}

As discussed in Section~\ref{sec:theory}, the performance EM algorithm with unknown weights does not depend on the ordering of the initialization means. We conjuncture that in general, this property that is a consequence of over-parameterization leads to the boost that is observed in the performance of EM with unknown weights.

We support this conjecture by revisiting the previous simulations with a different way of running EM for Model 1. For each set of $k$ vectors selected to be used as initial component means, we run EM $k!$ times, each using a different one-to-one assignment of these vectors to initial component means. We measure the empirical success probability $P_3$ based on the \emph{lowest} observed error among these $k!$ runs of EM. The results are presented in Table~\ref{tab:tildeP} in Appendix~\ref{sec:tables}. In general, we observe $P_3\gtrsim P_2$ for all cases we have studied, which supports our conjecture. However, this procedure is generally more time-consuming than EM for Model 2 since $k!$ executions of EM are required.

\begin{table}
\begin{center}
\textbf{Success probabilities for mixtures of two Gaussians} (Section~\ref{sec:twoG}) \\
\begin{tabular}{|c|c||c|c|c|}
  \hline
  Separation & Sample size
  & $w^*_1=0.52$  & $w^*_1=0.7$ & $w^*_1=0.9$  \\
  \hline 
  \hline 
  \multirow{2}*{$\theta^*_2-\theta^*_1=2$} &
  $n=1000$ &
  \spacefrac{\textcolor{red}{0.799}}{\textcolor{blue}{0.500}} &
  \spacefrac{\textcolor{red}{0.497}}{\textcolor{blue}{0.800}} &
  \spacefrac{\textcolor{red}{0.499}}{\textcolor{blue}{0.899}}
  \\
  \cline{2-5}
  &
  $n=\infty$ &
  \spacefrac{\textcolor{red}{0.504}}{\textcolor{blue}{1.000}} &
  \spacefrac{\textcolor{red}{0.514}}{\textcolor{blue}{1.000}} &
  \spacefrac{\textcolor{red}{0.506}}{\textcolor{blue}{1.000}}
  \\
  \hline
\end{tabular}\\
\vspace{0.2cm}
\textbf{Success probabilities for mixtures of three or four Gaussians} (Section~\ref{sec:3G}) \\
\begin{tabular}{|c|c|c|c|}
\hline
 Case 1 & Case 2 & Case 3  &Case 4 \\
  \hline 
  \hline 
  \spacefrac{\textcolor{red}{0.164}}{\textcolor{blue}{0.900}} &
  \spacefrac{\textcolor{red}{0.167}}{\textcolor{blue}{1.000}} &
  \spacefrac{\textcolor{red}{0.145}}{\textcolor{blue}{0.956}} &
  \spacefrac{\textcolor{red}{0.159}}{\textcolor{blue}{0.861}}
  \\
\hline
\end{tabular}
\end{center}
  \caption{Success probabilities for EM on Model 1 and Model 2 (denoted $P_1$ and $P_2$, respectively), reported as
  $\spacefrac{\textcolor{red}{P_1}}{\textcolor{blue}{P_2}}$.%
  }
\label{tab:1}
\end{table}

\subsubsection*{Acknowledgements}

DH and JX were partially supported by NSF awards DMREF-1534910 and CCF-1740833,
and JX was also partially supported by a Cheung-Kong Graduate School of Business Fellowship.
We thank Jiantao Jiao for a helpful discussion about this problem.

\newpage
\bibliography{paper}
\bibliographystyle{plainnat}
\newpage
\appendix
\section{Proof of Theorem \ref{thm:Model1}}\label{sec:proofthm2}
Let us define $h(\theta,w^*_1):=H(\theta;\theta^*,w^*_1)$. First, it is straightforward to show that 
$$h(0,0.5)=0,$$ 
and
\begin{itemize}
\item $h(\theta,0.5)$ is concave for $\theta\geq 0$ and $h(\theta^*,0.5)=\theta^*$.
\item $h(\theta,0.5)$ is convex for $\theta\leq 0$ and $h(-\theta^*,0.5)=-\theta^*$.
\end{itemize} 
Hence, we have
\begin{eqnarray}
h(\theta,0.5)-\theta&=&\left\{\begin{aligned}
&>0, && \theta\in (-\infty,-\theta^*)\bigcup (0,\theta^*)\\
&=0, && \theta=-\theta^*,0,\theta^*\\
&<0, && \theta\in (-\theta^*,0)\bigcup (\theta^*,\infty) 
\end{aligned}\right.\label{eq:shapeH}
\end{eqnarray}
Therefore, if we can show that the curve of $h(\theta,w^*_1)$ is strictly above the curve $h(\theta,0.5)$ for all $w^*_1>0.5$ and $\theta<\theta^*$, i.e.,
\begin{eqnarray}
h(\theta,w^*_1)&>&h(\theta,0.5), \quad \quad \forall  w^*_1>0.5,\theta<\theta^*, \label{eq:H_eq0}
\end{eqnarray}
then by \eqref{eq:shapeH}, we have
\begin{eqnarray}
h(\theta,w^*_1)-\theta&>&h(\theta,0.5)-\theta\ \geq \ 0, \quad \quad \forall w^*_1>0.5, \theta\leq -\theta^*.\label{eq:H_eq1}
\end{eqnarray}
Further, since $h$ is continuous, we know there exists $\delta>0$ and $\theta_{\delta}$, such that
$$h(\theta_{\delta},w^*_1)\ <\ \theta_{\delta}, \quad \forall w^*_1\in [0.5,0.5+\delta].$$ 
Hence, with \eqref{eq:H_eq1} and continuity of function $h(\theta,w^*_1)-\theta$, we know for each $w^*_1\in (0.5,0.5+\delta]$, there exists $\theta_w\in (-\theta^*,0)$ (the smallest fixed point) such that   
\begin{eqnarray}
h(\theta_w,w^*_1)&=&\theta_w \quad \text{and} \quad h(\theta,w^*_1)\ >\ \theta, \quad \quad \forall \theta\in (-\infty,\theta_w).\n
\end{eqnarray}
Therefore, if we initialize $\pt{0}\leq -\theta^*$, the EM estimate will converge to $\theta_w$. Hence, our final step is to show \eqref{eq:H_eq1} which is proved in the following lemma:
\begin{lemma}[Proved in Appendix~\ref{sec:proofHeq}]\label{lem:Heq}
For all $w^*_1\neq 0.5$, we have
\begin{eqnarray}
h(\theta,w^*_1)&>&h(\theta,0.5), \quad \quad \forall \theta<\theta^*,\label{eq:Heq_eq1}
\end{eqnarray}
and for all $w^*_1\in [0,1]$, we have
\begin{eqnarray}
0\kleq \frac{\partial h(\theta,w^*_1)}{\partial \theta}\kleq e^{-\frac{(\theta^*)^2}{2}}\ <\ 1, \quad \quad \forall \theta\geq \theta^*. \label{eq:Heq_eq2}
\end{eqnarray}
\end{lemma}

In fact, by Lemma~\ref{lem:Heq}, \eqref{eq:shapeH} and the fact $h(\theta^*,w)\equiv \theta^*$, it is straightforward to show the following corollary
\begin{corollary}\label{lem:Hfixpoint}
For all $w^*_1\in [0,1]$, $h(\theta,w^*_1)$ has only one fixed point (a stable fixed point) in $(0,\infty)$, which is $\theta=\theta^*$.
\end{corollary}

\section{Proof of Theorem \ref{thm:main}}\label{sec:app_main}
From the discussion in Section~\ref{sec:theory}, we just need to prove Theorem \ref{thm:main} for $w^*_1>0.5$. We use the following the strategy to prove Theorem \ref{thm:main}.
\begin{enumerate}
\item Prove Lemma \ref{lem:initialarea} (see Section \ref{sec:mainproof}) and therefore WLOG, we can safely assume $\dotp{\vpt{t},\vtheta^*}>0$ and $\pw{t}>0.5$ for all  $t>0$.
\item Prove Theorem \ref{thm:main} when the mean parameters $\theta^*_i$ is in one dimension.
\item Show that we can reduce the multi-dimensional problem into the one dimensional one.
\item Show geometric convergence by proving an attraction basin around $(\vtheta^*,w^*_1)$.
\end{enumerate}
Each one of the steps is proved in the following subsections in order.

\subsection{Proof of Lemma \ref{lem:initialarea}}\label{sec:proofinitialarea}
First it is clear that $\pw{t}_1\in (0,1)$. Hence, due to our initialization setting $\pw{0}_1=\pw{0}_2=0.5$, we just need to show 
\begin{itemize}
\item For all $\dotp{\vtheta,\vtheta^*}>0, w_1\in [0.5,1)$, we have
\begin{eqnarray}
\dotp{G_{\theta}(\vtheta,w_1;\vtheta^*,w^*_1),\vtheta^*} \> 0 \kand G_{w}(\vtheta,w_1;\vtheta^*,w^*_1)\> 0.5. \label{eq:initial_eq1}
\end{eqnarray}
\item For all $\dotp{\vtheta,\vtheta^*}<0, w_1\in (0,0.5]$, we have
\begin{eqnarray}
\dotp{G_{\theta}(\vtheta,w_1;\vtheta^*,w^*_1),\vtheta^*} \< 0 \kand G_{w}(\vtheta,w_1;\vtheta^*,w^*_1)\< 0.5. \label{eq:initial_eq2}
\end{eqnarray}
\end{itemize}
and then by a simple induction argument, it is straightforward to show Lemma \ref{lem:initialarea} holds. Moreover, let $w_2=1-w_1$ and note that the symmetric property of $G_{\theta}$ and $G_{w}$, i.e.,
\begin{eqnarray}
G_{\theta}(\vtheta,w_1;\vtheta^*,w^*_1)+G_{\theta}(-\vtheta,w_2;\vtheta^*,w^*_1)&=&0\n\\
G_{w}(\vtheta,w_1;\vtheta^*,w^*_1)+G_{w}(-\vtheta,w_2;\vtheta^*,w^*_1)&=&1.\n
\end{eqnarray}
Hence, we just need to show \eqref{eq:initial_eq1} holds. Since for any orthogonal matrices $\vV$, we have 
\begin{eqnarray}
\dotp{ G_{\theta}(\vtheta,w_1;\vtheta^*,w^*_1),\vtheta^*} &=& \dotp{G_{\theta}(\vV\vtheta,w_1;\vV\vtheta^*,w^*_1),\vV\theta^*}\n\\
G_{w}(\vtheta,w_1;\vtheta^*,w^*_1) &=& G_{w}(\vV\vtheta,w_1;\vV\vtheta^*,w^*_1)\n
\end{eqnarray}
Hence, the claim made in \eqref{eq:initial_eq1} and \eqref{eq:initial_eq2} is invariant to rotation of the coordinates. Hence, WLOG, we assume that $\vtheta=(\|\vtheta\|,0,0,\ldots,0)$ and $\vtheta^*=(\theta^*_{\parallel},\theta^*_{\perp},0,\ldots,0)$ with $\theta^*_{\parallel}>0$. Let us first show $G_{w}(\vtheta,w;\vtheta^*,w^*_1)\> 0.5$. It is straightforward to show that
\begin{eqnarray}
G_{w}(\vtheta,w_1;\vtheta^*,w^*_1)&=&\int \frac{w_1 e^{y\|\theta\|}}{w_1 e^{y\|\theta\|}+w_2 e^{-y\|\theta\|}}\left(w^*_1\phi(y-\theta^*_{\parallel})+w^*_2\phi(y+\theta^*_{\parallel})\right)\dif y\n\\
&=:& g_w(\|\theta\|,w_1;\theta^*_{\parallel},w^*_1),\n
\end{eqnarray}
where $\phi(x)$ denotes the pdf for $d'-$dimensional standard Gaussian if $x\in \bbR^{d'}$. Hence, we just need to show that 
\begin{eqnarray}
g_w(\theta,w_1;\theta^*,w^*_1) \> 0.5, \quad \forall w_1\in [0.5,1), w^*_1\in (0.5,1), \theta>0, \theta^*>0.\label{eq:initial_eq3}
\end{eqnarray}
Note that
\begin{eqnarray}
\frac{\partial g_w(\theta,w_1;\theta^*,w^*_1)}{\partial w_1}&=&\int \frac{1}{\left(w_1e^{y\theta}+w_2e^{-y\theta}\right)^2} \left(w^*_1\phi(y-\theta^*)+w^*_2\phi(y+\theta^*)\right) \dif y\> 0.\n
\end{eqnarray}
Hence, we just need to show $g_w(\theta,0.5;\theta^*,w^*_1)> 0.5$. Note that
\begin{eqnarray}
g_w(\theta, 0.5;\theta^*,w^*_1)-0.5&=&\int \frac{e^{y\theta}}{e^{y\theta}+e^{-y\theta}} \left(w^*_1\phi(y-\theta^*)+w^*_2 \phi(y+\theta^*)\right) \dif y -0.5\n\\
&=&\int \frac{e^{y\theta}-e^{-y\theta}}{2(e^{y\theta}+e^{-y\theta})} \left(w^*_1\phi(y-\theta^*)+w^*_2 \phi(y+\theta^*)\right) \dif y \n\\
&=&\int_{y\geq 0} \phi(y)e^{-\frac{(\theta^*)^2}{2}}\cdot \left(\frac{(2w^*_1-1)\left(\cosh_y(\theta^*+\theta)-\cosh_y(\theta^*-\theta)\right)}{2\cosh_y(\theta)}\right)\dif y\n\\
&>& 0,\n
\end{eqnarray}
where $\cosh_y(x)=\frac{1}{2}(e^{yx}+e^{-yx})$. Hence, \eqref{eq:initial_eq3} holds. Now we just need to show $\dotp{G_{\theta}(\vtheta,w_1;\vtheta^*,w^*_1),\vtheta^*}> 0$. It is straightforward to show that all components of $G_{\theta}(\vtheta,w_1;\vtheta^*,w^*_1)$ are $0$ except for the first two components denoted as $\tilde{\theta}_1$ and $\tilde{\theta}_2$. For the second component $\tilde{\theta}_2$, we have
\begin{eqnarray}
\tilde{\theta}_2&=&\theta^*_{\perp}\int \frac{w_1 e^{y\|\theta\|}-w_2e^{-y\|\theta\|}}{w_1 e^{y\|\theta\|}+w_2 e^{-y\|\theta\|}}\left(w^*_1\phi(y-\theta^*_{\parallel})-w^*_2\phi(y+\theta^*_{\parallel})\right)\dif y\n\\
&=:& \theta^*_{\perp} \cdot s(\|\theta\|,w_1;\theta^*_{\parallel},w^*_1), \label{eq:s}
\end{eqnarray}
and for the first component $\tilde{\theta}_1$, we have
\begin{eqnarray}
\tilde{\theta}_1&=&\theta^*_{\parallel}\int \frac{w_1 e^{y\|\theta\|}-w_2e^{-y\|\theta\|}}{w_1 e^{y\|\theta\|}+w_2 e^{-y\|\theta\|}}\left(w^*_1\phi(y-\theta^*_{\parallel})-w^*_2\phi(y+\theta^*_{\parallel})\right)\dif y\n\\
&&+\int \frac{w_1 e^{y\|\theta\|}-w_2e^{-y\|\theta\|}}{w_1 e^{y\|\theta\|}+w_2 e^{-y\|\theta\|}}\left(w^*_1(y-\theta^*_{\parallel})\phi(y-\theta^*_{\parallel})+w^*_2(y+\theta^*_{\parallel})\phi(y+\theta^*_{\parallel})\right)\dif y\n\\
&\stackrel{(a)}{=}& \theta^*_{\parallel} \cdot s(\|\theta\|,w_1;\theta^*_{\parallel},w^*_1)+\|\theta\|\int \frac{4w_1w_2}{\left(w_1 e^{y\|\theta\|}+w_2 e^{-y\|\theta\|}\right)^2}\left(w^*_1\phi(y-\theta^*_{\parallel})+w^*_2\phi(y+\theta^*_{\parallel})\right)\dif y\n\\
&>& \theta^*_{\parallel} \cdot s(\|\theta\|,w_1;\theta^*_{\parallel},w^*_1), \label{eq:initial_eq4}
\end{eqnarray}
where equation (a) holds due to partial integration. Hence, by \eqref{eq:s} and \eqref{eq:initial_eq4} and $\theta^*_{\parallel}>0$, we have
\begin{eqnarray}
\dotp{G_{\theta}(\vtheta,w_1;\vtheta^*,w^*_1),\vtheta^*}&> &\|\theta^*\|^2 \cdot s(\|\theta\|,w_1;\theta^*_{\parallel},w^*_1).\n
\end{eqnarray}
Hence, we just need to show 
\begin{eqnarray}
s(\theta,w_1;\theta^*,w^*_1)&>&0, \quad \forall \theta>0, w_1\in [0.5,1], \theta^*>0, w^*_1\in (0.5,1). \label{eq:initial_eq5}
\end{eqnarray}
For $w_1=0.5$, by \eqref{eq:s}, we have
\begin{eqnarray}
s(\theta,0.5;\theta^*,w^*_1)&=&\int \frac{e^{y\theta}-e^{-y\theta}}{e^{y\theta}+e^{-y\theta}}\left(w^*_1\phi(y-\theta^*)-w^*_2\phi(y+\theta^*)\right)\dif y\n\\
&=&\int_{y\geq 0} \frac{e^{y\theta}-e^{-y\theta}}{e^{y\theta}+e^{-y\theta}}\phi(y)e^{-\frac{(\theta^*)^2}{2}}\left(e^{y\theta^*}-e^{-y\theta^*}\right)\dif y \>0 .\label{eq:initial_eq6}
\end{eqnarray} 
For $w_1\in (0.5,1]$, by \eqref{eq:s} and taking derivative with respect to $w^*_1$, we have
\begin{eqnarray}
\frac{\partial s(\theta,w_1;\theta^*,w^*_1)}{\partial w^*_1}&=&\int \frac{w_1 e^{y\theta}-w_2e^{-y\theta}}{w_1 e^{y\theta}+w_2 e^{-y\theta}}\left(\phi(y-\theta^*)+\phi(y+\theta^*)\right)\dif y\n\\
&=&\int_{y\geq 0} \frac{2(w_1^2-w^2_2)}{\left(w_1 e^{y\theta}+w_2 e^{-y\theta}\right)\left(w_1 e^{-y\theta}+w_2 e^{y\theta}\right)}\left(\phi(y-\theta^*)+\phi(y+\theta^*)\right)\dif y\n\\
&>&0.\n
\end{eqnarray}
Hence, we just need to show
\begin{eqnarray}
s(\theta,w_1;\theta^*,0.5)&\geq&0, \quad \forall \theta>0, w_1\in (0.5,1], \theta^*>0. \label{eq:initial_eq7}
\end{eqnarray}
Note that
\begin{eqnarray}
2s(\theta,w_1;\theta^*,0.5)&=&\int \frac{w_1 e^{y\theta}-w_2e^{-y\theta}}{w_1 e^{y\theta}+w_2 e^{-y\theta}}\left(\phi(y-\theta^*)-\phi(y+\theta^*)\right)\dif y\n\\
&=&\int_{y\geq 0} \frac{w_1w_2 (e^{2y\theta}-e^{-2y\theta})}{\left(w_1 e^{y\theta}+w_2 e^{-y\theta}\right)\left(w_1 e^{-y\theta}+w_2 e^{y\theta}\right)}\left(\phi(y-\theta^*)-\phi(y+\theta^*)\right)\dif y\n\\
&\geq&0.\n
\end{eqnarray}
Hence, we have \eqref{eq:initial_eq7} holds. Combine with \eqref{eq:initial_eq6}, we have \eqref{eq:initial_eq5} holds which completes the proof of this lemma.

\subsection{Proof of Theorem \ref{thm:main} in one dimension}
We filled out the proofs that have left out in Section \ref{sec:mainproof}, namely Lemma \ref{lem:converge}, Lemma \ref{lem:shapeg} and C.2c.

\subsubsection{Proof of Lemma \ref{lem:converge}}\label{sec:proofconverge}
Based on $(\theta_{\star},w_{\star})$, we divide the region of $S-\{(\theta_{\star},w_{\star})\}$ into 8 pieces:
\begin{itemize}
\item $R_1=\{(\theta,w)\in S: \theta \in [\theta_{\star},\min\{r(a_w),b_{\theta}\}), w\in (a_w,w_{\star}]\}-\{(\theta_{\star},w_{\star})\}$. 
\item $R_2=\{(\theta,w)\in S: \theta \in [\theta_{\star},\min\{r(a_w),b_{\theta}\}), w\in [w_{\star},b_w)\}-\{(\theta_{\star},w_{\star})\}$.
\item $R_3=\{(\theta,w)\in S: \theta \in (\max\{r(b_w),a_{\theta}\},\theta_{\star}], w\in (a_w,w_{\star}]\}-\{(\theta_{\star},w_{\star})\}$.
\item $R_4=\{(\theta,w)\in S: \theta \in (\max\{r(b_w),a_{\theta}\},\theta_{\star}], w\in [w_{\star},b_w)\}-\{(\theta_{\star},w_{\star})\}$.
\item $R_5=\{(\theta,w)\in S: \theta\leq r(b_w),w\in (a_w,w_{\star}]\}$.
\item $R_6=\{(\theta,w)\in S: \theta\leq r(b_w),w\in [w_{\star},b_w)\}$.
\item $R_7=\{(\theta,w)\in S: \theta\geq r(a_w),w\in (a_w,w_{\star}]\}$.
\item $R_8=\{(\theta,w)\in S: \theta\geq r(a_w),w\in [w_{\star},b_w)\}$.
\end{itemize}
Note that region $R_5$ to $R_8$ may not exists depending on the range of $r(w)$. Next, due to C.2a, we know the reference curve only crosses region $R_1$ and $R_4$. Note that $r^{-1}(\theta)$ exists on the regions $R_1, R_2,R_3$ and $R_4$. Hence, based on the points are above or below the reference curve $r$, we can further divide the region $R_1$ and $R_4$ into 4 pieces:
\begin{itemize}
\item $R_{11}=\{(\theta,w)\in R_1:   r^{-1}(\theta)\leq w\}$. 
\item $R_{12}=\{(\theta,w)\in R_1: r^{-1}(\theta)\geq w\}$.
\item $R_{41}=\{(\theta,w)\in R_4: w\leq r^{-1}(\theta)\}$.
\item $R_{42}=\{(\theta,w)\in R_4: w\geq r^{-1}(\theta)\}$.
\end{itemize}

Now let's define $m:S\rightarrow [0,\infty)$ based on the following 10 regions 
$$\{R_{11}, R_{12},  R_2, R_3, R_{41}, R_{42}, R_{5}, R_{6},R_7,R_8\}:$$
\begin{itemize}
\item If $(\theta,w)\in R_{11}$, $m(\theta,w)=(w_{\star}-w)(r(w)-\theta_{\star})$, which is the area of the rectangle $D(\theta,w)$ given by $(\theta_{\star},w_{\star}),(r(w),w)$.
\item If $(\theta,w)\in R_{12}$, $m(\theta,w)=(w_{\star}-r^{-1}(\theta))(\theta-\theta_{\star})$, which is the area of the rectangle $D(\theta,w)$ given by $(\theta_{\star},w_{\star}),(\theta,r^{-1}(\theta))$.
\item If $(\theta,w)\in R_2$, $m(\theta,w)=(w-r^{-1}(\theta))(\theta-r(w))$, which is the area of the rectangle $D(\theta,w)$ given by $(r(w),r^{-1}(\theta)),(\theta,w)$.
\item If $(\theta,w)\in R_{3}$, $m(\theta,w)=(r^{-1}(\theta)-w)(r(w)-\theta)$, which is the area of the rectangle $D(\theta,w)$ given by $(r(w),r^{-1}(\theta)),(\theta,w)$.
\item If $(\theta,w)\in R_{41}$, $m(\theta,w)=(r^{-1}(\theta)-w_{\star})(\theta_{\star}-\theta)$, which is the area of the rectangle $D(\theta,w)$ given by $(\theta_{\star},w_{\star}),(\theta,r^{-1}(\theta))$.
\item If $(\theta,w)\in R_{42}$, $m(\theta,w)=(w-w_{\star})(\theta_{\star}-r(w))$, which is the area of the rectangle $D(\theta,w)$ given by $(\theta_{\star},w_{\star}),(r(w),w)$.
\item If $(\theta,w)\in R_5$, $m(\theta,w)=(b_w-w)(r(w)-\theta)$, which is the area of the rectangle $D(\theta,w)$ given by $(r(w),b_w),(\theta,w)$.
\item If $(\theta,w)\in R_6$, $m(\theta,w)=(b_w-w_{\star})(\theta_{\star}-\theta)$, which is the area of the rectangle $D(\theta,w)$ given by $(\theta,b_w),(\theta_{\star},w_{\star})$.
\item If $(\theta,w)\in R_7$, $m(\theta,w)=(w_{\star}-a_w)(\theta-\theta_{\star})$, which is the area of the rectangle $D(\theta,w)$ given by $(\theta_{\star},w_{\star}),(\theta,a_w)$.
\item If $(\theta,w)\in R_8$, $m(\theta,w)=(w-a_w)(\theta-r(w))$, which is the area of the rectangle $D(\theta,w)$ given by $(r(w),a_w),(\theta,w)$.
\end{itemize}

It is straightforward to show that function $m$ is a continuous function by checking the boundary and continuity of the reference function $r$. Further, $(\theta_{\star},w_{\star})$ is indeed the only solution for $m(\theta,w)=0$. Moreover, our construction of the rectangle $D$ makes sure that 
\begin{eqnarray}
\text{If~} (\tilde{\theta},\tilde{w}) \text{~is strictly inside~} D(\theta,w), \ \text{~then~} D(\tilde{\theta},\tilde{w})\subsetneq D(\theta,w).\label{eq:strictcontain}
\end{eqnarray} 
Next, we shall discuss the movement of the iterates from point $(\pt{t},\pw{t})$ to point $(\pt{t+1},\pw{t+1})$. For a given $\pw{t} \in[a_w,b_w]$, consider all the fixed points $\mathcal{V}$ in $[a_{\theta},b_{\theta}]$ for $g_{\theta}(\theta,w)$ with respect to $\theta$. Then, for any $\pt{t}\in (a_{\theta},b_{\theta})$, it should be inside an interval defined by $[q_1,q_2]$ where $q_1,q_2\in \mathcal{V}\bigcup \{a_{\theta},b_{\theta}\}$ and at least one of $q_1$ or $q_2$ is either a stable fixed point or one of $a_{\theta},b_{\theta}$. Further, since $g_{\theta}(\theta,w)$ is a non-decreasing function of $\theta$ and $(\pt{t+1},\pw{t+1})\in S$, we know $\pt{t+1}=g_{\theta}(\pt{t},\pw{t})\in [q_1,q_2]$ as well. Hence, comparing to the previous iteration $\pt{t}$, $\pt{t+1}=g_{\theta}(\pt{t},\pw{t})$ should (i) stay at a fixed point, i.e., $q_1$ or $q_2$ or (ii) move towards a stable fixed point $q_i$ or $a_{\theta},b_{\theta}$. Further, if $\pt{t+1}$ moves towards $a_{\theta}$ or $b_{\theta}$, then $a_{\theta}$ or $b_{\theta}$ has to be a stable fixed point as well. In other words, suppose $\pt{t+1}$ move towards $a_{\theta}$ and $a_{\theta}$ is not a stable fixed point. Then $a_{\theta}$ is not a fixed point as well and there exists a constant $c>0$ such that $\lim_{\theta\rightarrow a_{\theta}}g_{\theta}(\theta,\pw{t})\leq a_{\theta}-c$. Hence by choosing $\theta$ close enough to $a_{\theta}$, we know $g_{\theta}(\theta,w)<a_{\theta}$ which contradicts C.1. Now, by C.2b, C.2c and discussing which region $(\theta,w)$ belongs to, we can prove 

\begin{eqnarray}
\text{Point~}(\pt{t+1},\pw{t+1}) \text{~is strictly inside~} D(\pt{t},\pw{t}) \kand m(\pt{t+1},\pw{t+1})&<&m(\pt{t},\pw{t}). \n\\
\label{eq:conv_eq0}
\end{eqnarray} 

and 

\begin{eqnarray}
\text{If~} (\pt{t},\pw{t})\in R_1\bigcup R_2\bigcup R_3 \bigcup R_4, \quad \text{then~} (\pt{t+1},\pw{t+1})\in R_1\bigcup R_2\bigcup R_3 \bigcup R_4.\label{eq:conv_eq00}
\end{eqnarray}

Note that depending on the regions, there are total 10 cases. But for simplicity, we show the proof for two cases: $R_{11}$ and $R_{6}$ and leave the rest of the cases to the readers. For the first example, if point $(\pt{t},\pw{t})\in R_{11}$, then we know there exists a fixed point $\theta_s\in [\theta_{\star},b_{\theta}]$ for $g_{\theta}$ and $w_s\in [a_w,w_{\star}]$ for $g_w$ such that $\pt{t+1}=g_{\theta}(\pt{t},\pw{t})$ lies in between $\pt{t}$ and $\theta_s$, and $\pw{t+1}=g_{w}(\pt{t},\pw{t})$ lies in between $\pw{t}$ and $w_s$. Hence $(\pt{t+1},\pw{t+1})$ can only stay in $R_{1}$ which proves \eqref{eq:conv_eq00} for the case $(\pt{t},\pw{t})\in R_{11}$. Further, we have

\begin{eqnarray}
|g_{\theta}(\pt{t},\pw{t})-\theta_s|&\leq&|\pt{t}-\theta_s|, \label{eq:conv_eq1} \\
|g_{w}(\pt{t},\pw{t})-w_s|&\leq& |\pw{t}-w_s|,\label{eq:conv_eq2}
\end{eqnarray} 

where equality \eqref{eq:conv_eq1}/\eqref{eq:conv_eq2} holds if and only if $\pt{t}=\theta_s$/$\pw{t}=w_s$. Hence, by C.2, we have 
\begin{itemize}
\item If $\pt{t}=\theta_{\star}$, then $\pw{t}<w_{\star}$. Hence we have $\theta_s\in (\theta_{\star}, r(\pw{t}))$ and $w_s=w_{\star}$. and therefore, \eqref{eq:conv_eq2} is strict inequality. Hence, $\pw{t}<\pw{t+1}$. 
\item If $\pt{t}>\theta_{\star}$, then $\max(\theta_s,\pt{t})\leq r(\pw{t})$ and $w_s>r^{-1}(\pt{t})\geq \pw{t}$, therefore,
\begin{eqnarray}
\pt{t+1}\=g_{\theta}(\pt{t},\pw{t})\leq r(\pw{t}), \kand \pw{t}<g_w(\pt{t},\pw{t})\=\pw{t+1}.\label{eq:conv_eq3}
\end{eqnarray}
\end{itemize}
Therefore point $(\pt{t+1},\pw{t+1})$ lies in the rectangle $D(\pt{t},\pw{t})$ no matter what. Further, due to monotonic property of function $r$, we have

\begin{eqnarray}
r(\pw{t})>r(g_w(\pt{t},\pw{t})).\label{eq:conv_eq4}
\end{eqnarray}

Hence, by \eqref{eq:conv_eq3} and \eqref{eq:conv_eq4}, no matter what region $R_{11}$ or $R_{12}$ contains the point $(\pt{t+1},\pw{t+1})$, the rectangle $D(\pt{t+1},\pw{t+1})$ is strictly smaller than the rectangle $D(\pt{t}, \pw{t})$. Hence, we have \eqref{eq:conv_eq0} holds for the case $(\pt{t},\pw{t})\in R_{11}$. For the second example that if $(\theta,w)\in R_{6}$, then by C.2, we know there exists a fixed point $\theta_s \in (r(b_w),\theta_{\star}]$ for $g_{\theta}$ and $w_s\in [w_{\star},b_w]$ for $g_w$ such that $\pt{t+1}=g_{\theta}(\pt{t},\pw{t})$ lies in between $\pt{t}$ and $\theta_s$; and $\pw{t+1}=g_{w}(\pt{t},\pw{t})$ lies in between $\pw{t}$ and $w_s$. Hence, point $(\pt{t+1},\pw{t+1})$ can only stay in the region $R_{6}$ or $R_4$. Further, we have
\begin{eqnarray}
| g_{\theta}(\pt{t},\pw{t})-\theta_s| &\leq& |\pt{t}-\theta_s|,\n
\end{eqnarray}
where equality holds if and only if $\pt{t}=\theta_s$. Therefore, we have 
$$\pt{t+1}\=g_{\theta}(\pt{t},\pw{t})\> \pt{t},$$
and hence, no matter what region $R_{6}$ or $R_{4}$ contains the point $(\pt{t+1},\pw{t+1})$, the rectangle $D(\pt{t+1},\pw{t+1})$ is strictly smaller than the rectangle $D(\pt{t}, \pw{t})$. Similarly, we can show \eqref{eq:conv_eq0} holds for all other cases. Next, we claim that if point $(\pt{0},\pw{0})\in R_5\bigcup R_6\bigcup R_7 \bigcup R_8$, then within finite steps $t_0$, the estimate $(\pt{t_0},\pw{t_0})$ should lie in the region $R_1\bigcup R_2\bigcup R_3 \bigcup R_4$. Suppose point $(\pt{0},\pw{0})\in R_6$, $g_{\theta}(\theta,w)/\theta$ is continuous on $[\pt{0},r(b_w)]\times [w_{\star},b_w]$. Further, due to \eqref{eq:conv_eq0}, we have
\begin{eqnarray}
g_{\theta}(\theta,w)/\theta&>&1, \quad \forall (\theta,w)\in [\pt{0},r(b_w)]\times [w_{\star},b_w].\n
\end{eqnarray}   
Therefore, there exists a constant $\rho>1$ such that $g_{\theta}(\theta,w)\geq \rho \theta$ on $[\pt{0},r(b_w)]\times [w_{\star},b_w]$. Hence, within finite steps, we have $(\pt{t_0},\pw{t_0})\in R_1\bigcup R_2\bigcup R_3 \bigcup R_4$. Similarly we can show for $(\pt{0},\pw{0})\in R_5,R_7,R_8$ as well. Hence, by \eqref{eq:conv_eq00}, we just need to focus on $(\pt{0},\pw{0})\in R_1\bigcup R_2\bigcup R_3\bigcup R_4$. Now we use contradiction to prove that $m(\pt{t},\pw{t})$ converges to $0$. Suppose $m(\pt{t},\pw{t})$ does not converge to $0$, then by definition of $m$, we know there exists some constant $c_{\theta}>0$ and $c_w>0$, such that

\begin{eqnarray}
|\theta_{\star}-\pt{t}|\geq c_{\theta} \kand |w_{\star}-\pw{t}|\geq c_{w}, \quad \forall t\geq 0.
\end{eqnarray}

Further, since $S\supset D(\pt{0},\pw{0})\supset D(\pt{1},\pw{1})\supset\cdots$, we know all points $(\pt{t},\pw{t})$ are bounded on a compact set $D(\pt{0},\pw{0})$. Now consider function
$$U(\pt{t},\pw{t}):=\frac{m(\pt{t+1},\pw{t+1})}{m(\pt{t},\pw{t})}$$
we know $U$ is continuous on $(\pt{t},\pw{t})\in Q=\{(\theta,w_1)\in D(\pt{0},\pw{0}):  |\theta_{\star}-\theta |\geq c_{\theta}, |w_{\star}-w|\geq c_{w}\}.$ Further, since $Q$ is a compact set and $U<1$ on $Q$, we know there exists constant $\rho<1$ such that $\sup_{Q}U(\theta,w)\leq \rho$. Hence, we have $m(\pt{t},\pw{t})$ converges to $0$. Therefore, $(\pt{t},\pw{t})$ converges to $(\theta_{\star},w_{\star})$ since it is the only solution for $m=0$ and $m$ is continuous.

\subsubsection{Proof of Lemma \ref{lem:shapeg}}\label{sec:proofshapeg}
We study the shape of $g_w$ by its first, second and third derivatives. Note that (with $w_2=1-w_1$)
\begin{eqnarray}
\frac{\partial g_w(\theta,w_1)}{\partial w_1}&=&\bbE_{y\sim f^*}\left[\frac{1}{\left(w_1e^{y\theta}+w_2e^{-y\theta}\right)^2}\right] \> 0 \label{eq:shapeg_eq1111}\\
\frac{\partial^2 g_w(\theta,w_1)}{\partial w_1^2}&=&\bbE_{y\sim f^*}\left[\frac{e^{-y\theta}-e^{y\theta}}{\left(w_1e^{y\theta}+w_2e^{-y\theta}\right)^3} \right] \label{eq:shapeg_eq2}\\
\frac{\partial^3 g_w(\theta,w_1)}{\partial w_1^3}&=&\bbE_{y\sim f^*}\left[\frac{\left(e^{y\theta}-e^{-y\theta}\right)^2}{\left(w_1e^{y\theta}+w_2e^{-y\theta}\right)^4}\right] \>0 \label{eq:shapeg_eq3}
\end{eqnarray}
Hence, by \eqref{eq:shapeg_eq3}, we know the second derivative $\frac{\partial^2 g_w(\theta,w_1)}{\partial w_1^2}$ is a strictly increasing function of $w_1$ if $\theta\neq 0$. Hence, the second derivative can only change the sign at most once, the shape of $g_w$ can only be one of the following three cases: (i) concave (the second derivative is always negative), (ii) concave-convex (the second derivative is negative, then positive) and (iii) convex (the second derivative is always positive). Note that by Lemma \ref{lem:initialarea}, we know $g_w(\theta,0.5)>0.5$ if $\theta>0$. Moreover, it is easy to check that $g(\theta,0)=0$ and $g(\theta,1)=1$. Hence, we know for $\theta>0$, the shape of $g_w$ can only be either case (i) or case (ii). For case (i), it is clear that we have $1$ is the only stable fixed point and 
\begin{eqnarray}
g_w(\theta,w_1)>w_1\quad \text{is equivalent to} \quad w_1\in (0,1).\label{eq:shapeg_eq1}
\end{eqnarray}
For case (ii), then depends on the value of the derivative at $w_1=1$ i.e., $\partial g_w(\theta,w_1)/\partial w_1|_{w_1=1}$, we have 
\begin{itemize}
\item If $\partial g_w(\theta,w_1)/\partial w_1|_{w_1=1}\leq 1$, $w_1=1$ is the stable fixed point and \eqref{eq:shapeg_eq1} holds. 
\item If $\partial g_w(\theta,w_1)/\partial w_1|_{w_1=1}<1$, then $w_1=1$ is only a fixed point and there exists a stable fixed point in $(0,1)$ such that \eqref{eq:shapeofg} holds.
\end{itemize}


\subsection{Proof of C.2b}\label{sec:proofc2bfirst}
According to \eqref{eq:r}, function $r$ is a one to one mapping between $w\in (0.5,1]$ and $\theta\in [(w^*_1-w^*_2)\theta^*,\infty)$. Hence, 
we can simplify C.2b as  
\begin{itemize}
\item If $w_1\in (w^*_1,1]$, then $w_1>w_s>w^*_1$,  
\item If $w_1=w^*_1$, then $w_1=w_s=w_{\star}$,  
\item If $w_1\in (0.5,w^*_1)$, then $w_1<w_s<w^*_1$,  
\end{itemize}  
where $w_s$ is any stable fixed point in $[a_w,b_w]$ or fixed point in $(a_w,b_w)$ for $\theta=r(w_1)$. By \eqref{eq:shapeofg} in Lemma \ref{lem:shapeg}, we can complete the proof for C.2b by showing the following technical lemma proved in Appendix~\ref{sec:proofc2b}:
\begin{lemma} \label{lem:c2b}
Let $\gamma=\frac{2w^*_1-1}{2w_1-1}$, we have
\begin{eqnarray}
g_w(\gamma \theta^*,w_1)&<&w_1 \kand g_w(\gamma \theta^*,w_1^*)>w_1^* \quad \forall w_1\in (w^*_1,1]\n\\
g_w(\gamma \theta^*,w_1)&>&w_1 \kand g_w(\gamma \theta^*,w_1^*)<w_1^* \quad \forall w_1\in (0.5,w_1^*)\n
\end{eqnarray}
\end{lemma}

\subsubsection{Proof of C.2c} \label{sec:proofc2c}
Recall our construction of the adjusted reference curve $r_{adj}$ in Section \ref{sec:mainproof}, we have
$$r_{adj}(w)\=r(w)-\epsilon\cdot \max(0,w-1+\delta) \= \frac{2w^*_1-1}{2w-1}\theta^*-\epsilon\cdot \max(0,w-1+\delta),$$
for some positive $\epsilon,\delta>0$. Also, note that $g_{\theta}(\theta,1)\equiv (2w^*_1-1)\theta^*$. Hence, we just need to show the following 
\begin{itemize}
\item[C.2c'] Given $w_1\in (a_w,b_w)$, any stable fixed point $\theta_s$ of $g_{\theta}(\theta,w)$ in $[a_{\theta},b_{\theta}]$ or fixed point $\theta_s$ in $(a_{\theta},b_{\theta})$ satisfies that
\begin{itemize}
\item If $w_1<w_{\star}$, then $r(w)> \theta_s>\theta_{\star}$. 
\item If $w_1=w_{\star}$, then $r(w)= \theta_s=\theta_{\star}$.   
\item If $w_1>w_{\star}$, then $r(w)< \theta_s<\theta_{\star}$.   
\end{itemize}
\end{itemize}

Like the proof for C.2b shown in Section \ref{sec:mainproof}, we first show that there exists stable fixed point for $g_{\theta}(\theta,w_1)$ with respect to $\theta$, i.e.,
\begin{itemize}
\item[Claim 1] If $w_1\in (0.5,w^*_1]$, then there exists an unique non-negative fixed point for $g_{\theta}(\theta,w_1)$ denoted as $F_{\theta}(w_1)$. Further, $F_{\theta}(w_1)\geq \theta^*$.
\item[Claim 2] If $w_1\in (w^*_1,1]$, then there exists positive stable fixed point for $g_{\theta}(\theta,w_1)$ and all non-negative fixed points are in $(0,\theta^*)$. 
\end{itemize}
First, it is clear that $\theta=0$ is not a fixed point for $w_1>0.5$ and $w^*_1>0.5$, therefore, we just need to consider $\theta>0$. Then, to prove Claim 1 and Claim 2, we should find out the shape of $g_{\theta}(\theta, w_1)$ for different true values $(\theta^*,w^*_1)$. Notice that, by Lemma \ref{lem:Heq}, we know the shape of $H(\theta,w_1;\theta^*)=G_{\theta}(\theta,w_1;\theta^*,w_1)$, i.e., for $\theta>0, w_1\in [0.5,1]$
\begin{eqnarray}
H(\theta,w_1;\theta^*) &\gtreqless& \theta \quad \text{is equivalent to} \quad \theta\lesseqgtr \theta^*.\label{eq:c2c_eq1}
\end{eqnarray}  
Hence, our next step to compare $G_{\theta}(\theta,w_1;\theta^*,w^*_1)$ with $H(\theta,w_1;\theta^*)=G_{\theta}(\theta,w_1;\theta^*,w_1)$. Note that, we have
\begin{eqnarray}
&&\frac{\partial G_{\theta}(\theta,w_1;\theta^*,w^*_1)}{\partial w^*_1}\= \int y \frac{w_1e^{y\theta}-w_2e^{-y\theta}}{w_1e^{y\theta}+w_2e^{-y\theta}} \left(\phi(y-\theta^*)-\phi(y+\theta^*)\right) \dif y\n\\
&=&\int_{y\geq 0}\left(\frac{w_1e^{y\theta}-w_2e^{-y\theta}}{w_1e^{y\theta}+w_2e^{-y\theta}} +\frac{w_1e^{-y\theta}-w_2e^{y\theta}}{w_1e^{-y\theta}+w_2e^{y\theta}} \right) y\left(\phi(y-\theta^*)-\phi(y+\theta^*)\right) \dif y\n\\
&=&2\int_{y\geq 0}\frac{w_1-w_2}{\left(w_1e^{y\theta}+w_2e^{-y\theta}\right)\left(w_1e^{-y\theta}+w_2e^{y\theta}\right)}y\left(\phi(y-\theta^*)-\phi(y+\theta^*)\right) \dif y \>0.\n\\
\label{eq:fwdif}
\end{eqnarray}
Hence, if $w_1\in (w^*_1,1]$, we know $G_{\theta}$ will be strictly below $H$. Therefore
\begin{eqnarray}
G_{\theta}(\theta,w_1;\theta^*,w^*_1)\<\theta, \quad \forall \theta\geq \theta^*.\n
\end{eqnarray}
Hence, with $G_{\theta}(0,w_1;\theta^*,w^*_1)=(w_1-w_2)(w_1^*-w_2^*)\theta^*>0$ and continuity of the function, we know Claim 2 holds. Similarly, if $w_1\in (0.5,w^*_1]$, we know $G_{\theta}$ will be strictly above $H$. Therefore
\begin{eqnarray}
G_{\theta}(\theta,w_1;\theta^*,w^*_1)\> \theta, \quad \forall 0<\theta\leq \theta^*.\n
\end{eqnarray}
Hence, to prove Claim 1, we just need to show that $G_{\theta}(\theta,w_1;\theta^*,w^*_1)$ is bounded by some constant $C$ and 
\begin{eqnarray}
\frac{\partial G_{\theta}(\theta,w_1;\theta^*,w^*_1)}{\partial \theta}&<& 1, \quad  \forall \theta\geq \theta^*,0.5<w_1\leq w^*_1\label{eq:fdif}.
\end{eqnarray}  
To prove boundedness, we have the following more general lemma:
\begin{lemma}[Proved in Appendix~\ref{sec:proofbound}]\label{lem:bound}
Given any $(\vtheta,w_1,\vtheta^*,w^*_1)$, we have
\[
	\|G_{\theta}(\vtheta,w_1;\vtheta^*,w^*_1)\|^2
	\kleq
	1+\|\vtheta^*\|^2
	.
\]
Hence, for all $t\geq 1$, $\|\vpt{t}\|^2\leq \|\vtheta^*\|^2+1.$
\end{lemma}
To prove \eqref{eq:fdif}, we have for $\theta\geq \theta^*$,
\begin{eqnarray}
\frac{\partial G_{\theta}(\theta,w_1;\theta^*,w^*_1)}{\partial \theta}
&=&\int\frac{4w_1w_2}{\left(w_1 e^{y\theta}+w_2e^{-y\theta}\right)^2}y^2\left(w^*_1\phi(y-\theta^*)+w^*_2\phi(y+\theta^*)\right)\dif y\n\\
&=&\frac{\partial H(\theta,w_1;\theta^*)}{\partial \theta}+(w^*_1-w_1)\int\frac{4w_1w_2}{\left(w_1e^{y\theta}+w_2e^{-y\theta}\right)^2}y^2\left(\phi(y-\theta^*)-\phi(y+\theta^*)\right)\dif y\n\\
&\stackrel{(i)}{\leq}&\frac{\partial H(\theta,w_1;\theta^*)}{\partial \theta}\n\\
&\stackrel{(ii)}{\leq}&e^{-\frac{(\theta^*)^2}{2}}<1\n,
\end{eqnarray}
where inequality (ii) holds due to Lemma \ref{lem:Heq} and inequality (i) holds due to
\begin{eqnarray}
w_1 e^{y\theta}+w_2 e^{-y\theta}&\geq&w_1 e^{-y\theta}+w_2 e^{y\theta},\quad \forall \theta>0.\n
\end{eqnarray}
This completes the proof for Claim 1 and Claim 2. Finally, it is straightforward to show the rest of C.2c by Claim 1 and Claim 2 and the following lemma: 
\begin{lemma}[Proved in Appendix~\ref{sec:proofc2c2c}]\label{lem:c2c2c}
\begin{eqnarray}
g_{\theta}(\gamma\theta^*,w_1)&<&\gamma\theta^*, \quad \forall w_1\in (\frac{1}{2},w_1)\label{eq:v3}\\
g_{\theta}(b\theta^*,w_1)&>& b\theta^*, \quad \forall b\in (0,\gamma], w_1\in (w_1,1).\label{eq:v4}
\end{eqnarray}
\end{lemma}

\subsection{Reduction to one dimension}\label{sec:reduceto1}
In this section, we show how to reduce multi-dimensional problem into one-dimensional problem by proving the angle between the two vectors $\vtheta^*$ and $\vpt{t}$ is decreasing to $0$. Define 
$$\pb{t}:=\arccos \frac{\dotp{\vpt{t},\vtheta^*}}{\|\vpt{t}\|\|\vtheta^*\|},$$ 
then given $\dotp{\vpt{0},\vtheta^*}>0$, we have
\begin{itemize} 
\item If $\pb{0}=0$, then for $t\geq 1$, we have $\pb{t}=0$, i.e., it is an one-dimensional problem.
\item If $\pb{0}\in (0,\frac{\pi}{2})$, then for $t\geq 1$, we have $\pb{t} \in (0,\pb{t-1})$.
\end{itemize}
We use similar strategy shown in \citep{xu2016global} to prove this. First let us define $\pa{t}:=\arccos \frac{\dotp{\vpt{t}, \vpt{t+1}}}{\|\vpt{t}\| \|\vpt{t+1}\|}$, i.e., the angle between the two vectors $\vpt{t}$ and $\vpt{t+1}$. Then since $\dotp{\vpt{0},\vtheta^*}>0$, we have $\pb{0}\in [0,\frac{\pi}{2})$. Further, it is straightforward to verify that if $\pb{0}=0$, we have $\pb{t}=0,\forall t\geq 0$. Hence, with Lemma \ref{lem:initialarea}, from now on, we assume $\pb{t}\in (0,\frac{\pi}{2})$ and $\pw{t}_1\in [0.5,1)$ for all $t\geq 0$. Therefore, we just need to show $\pb{t}<\pb{t-1}, \forall t>0$. To prove this, we just need to to prove the following three statements hold for
$\forall t\geq 0$:
\begin{enumerate}
  \item[(i)] $\pb{t}\in (0, \frac{\pi}{2})$.
  \item[(ii)]$\pa{t}\in (0, \pb{t})$.
  \item[(iii)]$\pb{t+1} = \pb{t}- \pa{t}\in (0, \pb{t})$.
\end{enumerate}
We use induction to show (i)-(iii) by proving the following chain of arguments:
\begin{description}
  \item[Claim 1] If (i) holds for $t$, then (ii) holds for $t$.
  \item[Claim 2] If (i) and (ii) hold for $t$, then (iii) holds for $t$.
  \item[Claim 3] If (i), (ii), and (iii) hold for $t$, then (i) holds for $t+1$.
\end{description}
Since (i) holds for $t=0$ and Claim 1 holds, it suffices to prove Claims 2-3. For simplicity, we drop $\dotp{t}$ in the notation and use $\tilde{\cdot}$ to indicate the values for the next iteration $t+1$, i.e., $\tilde{\vtheta}=\vpt{t+1}$ and $\tilde{\beta}=\pb{t+1}$. Since for any orthogonal matrix $\vV$, we have
\begin{eqnarray}
\vV G_{\theta}(\vtheta,w_1;\vtheta^*,w^*_1),\vtheta^* &=& G_{\theta}(\vV\vtheta,w_1;\vV\vtheta^*,w^*_1)\n\\
G_{w}(\vtheta,w_1;\vtheta^*,w^*_1) &=& G_{w}(\vV\vtheta,w_1;\vV\vtheta^*,w^*_1)\label{eq:rotate}
\end{eqnarray}
Hence, it is straightforward to check that the Claims are invariant under any rotation of the coordinates. Hence, WLOG, we assume that $\vtheta=(\|\vtheta\|,0,0,\ldots,0)$ and $\vtheta^*=(\theta^*_{\parallel},\theta^*_{\perp},0,\ldots,0)$ with $\theta^*_{\parallel}>0$ and $|\theta^*_{\perp}|>0$. Then, it is straightforward to show that all components of $\tilde{\vtheta}$ are $0$ except for the first two components denoted as $\tilde{\theta}_1$ and $\tilde{\theta}_2$. Hence, we just need to focus on the two-dimensional space spanned by the first two components. From \eqref{eq:s}, \eqref{eq:initial_eq4} and \eqref{eq:initial_eq5}, we have $\tan \alpha <\tan \beta=|\theta_{\perp}|/\theta_{\parallel}$ which implies Claim 2, and $\tilde{\theta}_2/\theta^*_{\perp}>0$ which implies Claim 3. Next, we want to prove the angle $\pb{t}$ is decreasing to $0$. Define $\ptpa{t}=\frac{|\dotp{\vpt{t},\vtheta^*}|}{\|\vpt{t}\|}$ and $\ptpe{t}=\|\vtheta^*-\ptpa{t}\|$. Hence, to show $\pb{t}$ decreases to $0$, it is equivalent to show that $\ptpa{t}$ converges to $\|\vtheta^*\|$. WLOG, we assume that $\vpt{0}=(\|\vpt{0}\|,0,0,\ldots,0)$ and $\vtheta^*=(\ptpa{0},\ptpe{0},0,\ldots,0)$ with $\ptpa{0}>0$ and $|\ptpe{0}|>0$. It is straightforward to show that the only non-zero components of $\vpt{t}$ are the first two components. Hence, we just need to analyze a two dimensional problem. Then, since $\pb{t}$ is decreasing, we have $\ptpa{t}=\|\vtheta^*\|\cdot \pb{t}$ is increasing. Hence 
\begin{eqnarray}
\ptpa{t}\in [\ptpa{1},\|\vtheta^*\|], \quad \forall t\geq 1.\label{eq:boundptpa}
\end{eqnarray}
To prove the increasing sequence $\ptpa{t+1}$ converges to $\|\vtheta^*\|$, we just need to show that for any $\hat{\theta}<\|\vtheta^*\|$, we can find $\ptpa{t+1}/\ptpa{t}\geq \rho_{\hat{\theta}}$ for some constant $\rho_{\hat{\theta}}>1$, then with a straightforward contradiction argument, within finite iterations, we should have $\ptpa{t'}>\hat{\theta}$ for a certain $t'$, which implies $\ptpa{t+1}$ converges to $\|\vtheta^*\|$. To find such $\rho$, note that, since $\ptpa{t}$ is a value invariant to coordinate rotations, by \eqref{eq:s},\eqref{eq:initial_eq4} and \eqref{eq:initial_eq5}, we have $U:=\ptpa{t+1}/\ptpa{t}$ is a continuous function of $\|\vpt{t}\|,\pw{t}_1$ and $\ptpa{t}$ and 
\begin{eqnarray}
\ptpa{t+1}/\ptpa{t}\> 1, \quad \forall \|\vpt{t}\|>0,\pw{t}_1\in (0.5,1], \ptpa{t}\in [\ptpa{1},\|\vtheta^*\|).\n
\end{eqnarray}

Hence, we just need to find some constants $0<c_1<c_2$ and $0.5<c_3<1$ such that $\|\vpt{t}\|\in [c_1,c_2]$ and $\pw{t}_1\in[c_3,1]$ for $t\geq 1$, then we can find $\rho$ by the uniform continuity argument. From Lemma \ref{lem:bound}, we have $c_2=1+\|\vtheta^*\|$. Since both $\|\vpt{t}\|$ and $\pw{t}_1$ is invariant to the coordinate rotations due to \eqref{eq:rotate}. WLOG, we assume that $\vpt{t}=(\|\vpt{t}\|,0)$ and $\vtheta^*=(\ptpa{t},\ptpe{t})$. Let us define the first coordinates of $\vpt{t+1}$ as $\tpt{t+1}_1$,  note that, we have
\begin{eqnarray}
\tpt{t+1}_1&=&\int y \frac{\pw{t}_1 e^{y\|\vpt{t}\|}-\pw{t}_2e^{-y\|\vpt{t}\|}}{\pw{t}_1 e^{y\|\vpt{t}\|}+\pw{t}_2 e^{-y\|\vpt{t}\|}}\left(w^*_1\phi(y-\ptpa{t})+w^*_2\phi(y+\ptpa{t})\right)\dif y\n\\
&=&G_{\theta}(\|\vpt{t}\|,\pw{t}_1;\ptpa{t},w^*_1)\n\\
\pw{t+1}_1&=&\int \frac{\pw{t}_1 e^{y\|\vpt{t}\|}-\pw{t}_2e^{-y\|\vpt{t}\|}}{\pw{t}_1 e^{y\|\vpt{t}\|}+\pw{t}_2 e^{-y\|\vpt{t}\|}}\left(w^*_1\phi(y-\ptpa{t})+w^*_2\phi(y+\ptpa{t})\right)\dif y\n\\
&=&G_{w}(\|\vpt{t}\|,\pw{t}_1;\ptpa{t},w^*_1)
\label{eq:mo_eq1}
\end{eqnarray}
Hence, $(\tpt{t+1}_1,\pw{t+1}_1)$ is the next iteration of $(\|\vpt{t}\|,\pw{t}_1)$ of the population-$\rm{EM}_{2}$ under the true value $(\ptpa{t},w^*_1)$. Indeed, we can consider this two dimensional problem as a series of one dimensional problems that follows this procedure:
\begin{itemize}
\item[Step 1] Start with point $(\|\vpt{1}\|,\pw{1}_1)\in S$, where $S=(0,\infty)\times (0.5,1)$.
\item[Step 2] For iteration $t$, let point $(\|\vpt{t}\|,\pw{t}_1)$ move towards the point $(\tpt{t+1}_1,\pw{t+1}_1)$ following the one dimensional update rule for the true value $\theta_{\star}=\ptpa{t}$.
\item[Step 3] Shift the true value $\theta_{\star}=\ptpa{t}$ and the point $(\tpt{t+1}_1,\pw{t+1}_1)$ to the right to their new values: true value $\theta_{\star}=\ptpa{t+1}$ and new point $(\|\vpt{t+1}\|,\pw{t+1}_1)$.
\item[Step 4] End iteration $t$ and go back to Step 2 for iteration $t+1$.
\end{itemize}

To analyze this, recall our analysis for the one dimension case in Section \ref{sec:mainproof}. Due to Lemma \ref{lem:shapeg} holds for any non-zero true value $\theta^*$, by typical uniform continuity argument, we can find $\delta,\epsilon>0$ such that the adjusted reference curve $r_{adj}(w_1;\theta_{\star})$ defined by 
$$r_{adj}(w_1;\theta_{\star})\= \frac{2w^*_1-1}{2w_1-1}\theta_{\star}-\epsilon \cdot \max(0,w_1+\delta-1)>0,$$
satisfies C.1,C.2 with $(a_{\theta},b_{\theta})=(0,\infty),(a_w,b_w)=(0.5,1)$ for any true value $\theta_{\star}\in [\ptpa{1},\|\vtheta^*\|]$ and $w_{\star}=w^*_1$. Hence, on $S=(0,\infty)\times (0.5,1)$, as $\theta_{\star}$ increases, the reference curve shifted to the right. Further, for any point $(\theta, w)$ in $S$, recall its corresponding area function $m(\theta,w)$ and rectangle $D(\theta,w)$ in the proof for Lemma~\ref{lem:converge} in Appendix~\ref{sec:proofconverge}. We use $m(\theta,w;\theta_{\star})$ and $D(\theta,w;\theta_{\star})$ to denote their values under the true value $\theta_{\star}$. By their definitions, we note that the left side and down side of the rectangle $D(\theta,w;\theta_{\star})$ is non-decreasing as $\theta_{\star}$ increases. Hence, by \eqref{eq:conv_eq0}, we know as $\ptpa{t}$ increases, $\pw{t}_1$ is always lower bounded by the down side of the rectangle $D(\|\vpt{1}\|,\pw{1}_1;\ptpa{1})$ due to the following chain of arguments:
\begin{eqnarray}
\pw{t+1}_1&\stackrel{(i)}{\geq}& \text{lower side of~} D(\|\vpt{t}\|,\pw{t}_1;\ptpa{t}) \ \stackrel{(ii)}{\geq} \ \text{lower side of~} D(\|\vpt{t}\|,\pw{t}_1;\ptpa{t-1})\n\\
&\stackrel{(iii)}{\geq}& \text{lower side of~} D(\tpt{t}_1,\pw{t-1}_1;\ptpa{t-1})\ \stackrel{(iv)}{\geq}\ \text{lower side of~} D(\|\vpt{t-1}\|,\pw{t-1}_1;\ptpa{t-1})\n\\ 
&\geq&\ \cdots \ \geq \ \text{lower side of~} D(\|\vpt{1}\|,\pw{1}_1;\ptpa{1})\:=\ c_3,\n
\end{eqnarray}   
where inequality (i) holds due to \eqref{eq:conv_eq0}, inequality (ii) and (iii) hold due to the shift of reference curve and definition of the rectangle $D$, and inequality (iv) holds due to \eqref{eq:strictcontain}. Also, we can show 
\begin{eqnarray}
\|\vpt{t}\| \geq \min\{\|\vpt{1}\|,(w^*_1-w^*_2)\ptpa{1}-\epsilon\delta\}:=c_1.\n
\end{eqnarray}
This is because,
\begin{itemize}
\item If $\|\vpt{t}\|\leq \ptpa{t}-\epsilon\delta$, i.e., point $(\|\vpt{t}\|,\pw{t}_1)$ is inside the region $R_5$ or $R_6$ defined by the true value $\theta_{\star}=\ptpa{t}$, then we know $\|\vpt{t+1}\|\geq \tpt{t+1}_1\geq \|\vpt{t}\|$.
\item If $\|\vpt{t}\|\leq \ptpa{t}-\epsilon\delta$, i.e., point $(\|\vpt{t}\|,\pw{t}_1)$ is inside the regions $R_1$-$R_4$ (note that regions $R_7$ and $R_8$ doesn't exists here), we have $(\tpt{t},\pw{t+1}_1)$  stay at $R_1$-$R_4$ and hence $\|\vpt{t+1}\|\geq \tpt{t+1}_1\geq  \ptpa{t}-\epsilon\delta$.
\end{itemize}
Hence, this completes the proof of our claim that the angle $\pb{t}$ is decreasing to $0$. Finally, we want to show that $(\|\vpt{t}\|,\pw{t}_1)$ converges to $(\|\vtheta^*\|,w^*_1)$ which implies $(\vpt{t},\pw{t}_1)$ converges to $(\vtheta^*,w^*_1)$ due to $\pb{t}\rightarrow 0$. To prove this final step, we just need to bound $\pw{t}_1$ away from $1$, i.e., there exists $c_4\in(0,1)$ such that
\begin{eqnarray}
\pw{t}_1\kleq c_4\<1, \quad \forall t\geq 1.\label{eq:wupper}
\end{eqnarray}
Note that if \eqref{eq:wupper} holds. Consider the following functions
 
\begin{eqnarray} 
U_1&=&m(\tpt{t+1}_1,\pw{t+1}_1;\ptpa{t})/m(\|\vpt{t}\|,\pw{t}_1;\ptpa{t})\n\\
U_2&=&m(\|\vpt{t+1}\|,\pw{t+1}_1;\|\vtheta^*\|)/m(\tpt{t+1}_1,\pw{t}_1;\ptpa{t})\n\\
U_3&=&m(\|\vpt{t}\|,\pw{t}_1;\|\ptpa{t}\|)/m(\|\vpt{t}\|,\pw{t}_1;\|\vtheta^*\|).\n
\end{eqnarray}

For any $\delta_0>0$, we have after finite iterations $t_1$, $\ptpa{t_1}$ will stay in the $\delta_0$-neighborhood around $\|\vtheta^*\|$. Hence, consider $t>t_1$, note that on the following compact set $S'$:
\begin{eqnarray}
	S'
	&:=&
	\left\{\pw{t}\in [c_3,c_4], \|\vpt{t}\|\in [c_1,c_2], \ptpa{t}\in [\|\vtheta^*\|-\delta_0,\|\vtheta^*\|]\right\}
		\n\\
	&&\quad \quad \quad \quad \quad \quad \quad \quad \quad \quad \quad \quad \quad \quad -\left\{(\|\vpt{t}\|-\|\vtheta^*\|)^2+(\pw{t}-w_1^*)^2<4\delta_0^2\right\}
	.	\n\\
		\label{eq:S'}
\end{eqnarray}   
we have $U_1<1$, therefore, we can find constant $\rho_1<1$ such that $U_1\leq \rho_1$ on $S'$. Further, we know there exists a constant $c'$ such that $\max(U_2,U_3)\leq (1+c\cdot\pb{t})$ on this compact set $S'$ since $\ptpa{t}=\cos\pb{t}\cdot \|\vtheta^*\|$ and $\tpt{t}=\cos\pb{t}\cdot \|\vpt{t}\|$. Hence for large enough $t_2$, there exists $\rho_2<1$ such that for any $t>t_2$ and point $(\|\vpt{t}\|,\pw{t}_1)$ in $S'$, we have
$$\frac{m(\|\vpt{t+1}\|,\pw{t+1}_1;\|\vtheta^*\|)}{m(\|\vpt{t}\|,\pw{t}_1;\|\vtheta^*\|)}\= U_1\cdot U_2\cdot U_3 \kleq \rho_2\<1. $$  
Hence, we have either $m(\|\vpt{t+1}\|,\pw{t+1}_1;\|\vtheta^*\|)$ is strictly decreasing at rate $\rho_2$ or $(\|\vpt{t}\|,\pw{t}_1)$ was in the $2\delta_0$-neighborhood around $(\|\vtheta\|^*,w^*_1)$ and therefore by the analysis in Lemma \ref{lem:converge}, there exists constant $c''>0$ and $c'''>0$ such that
$$m(\|\vpt{t+1}\|,\pw{t+1}_1;\|\vtheta^*\|)<(1+c''\cdot \pb{t})\cdot c'''\delta^2_0.$$ 
Either way, by arbitrary choice of $\delta_0$, we know $m(\|\vpt{t+1}\|,\pw{t+1}_1;\|\vtheta^*\|)$ converges to $0$ which implies $\vpt{t}$ converges to $\vtheta^*$. Hence, finally, we just need to bound $\pw{t}_1$. Note that in the proof of Lemma \ref{lem:converge}, we used the following strategy to show that $\pw{t}_1$ is bounded away from $1$:
\begin{itemize}
\item If $(\pt{0},\pw{0}_1)\in R_5\bigcup R_6$, within finite iterations $t_0$, $(\pt{t_0},\pw{t_0}_1)$ will reach the region $R_1\bigcup R_2 \bigcup R_3 \bigcup R_4$.
\item When $(\pt{t_0},\pw{t_0}_1)\in R_1\bigcup R_2 \bigcup R_3 \bigcup R_4$, by \eqref{eq:strictcontain} and \eqref{eq:conv_eq0}, we have for all $t\geq t_0$,
\begin{eqnarray}
(\pt{t+1},\pw{t+1}_1)\in D(\pt{t+1},\pw{t+1}_1)\stackrel{(a)}{\subseteq} D(\pt{t},\pw{t}_1)\subseteq \cdots \subseteq D(\pt{t_0},\pw{t_0}_1).\label{eq:wbound_eq1}
\end{eqnarray}
Hence, $\pw{t}\leq \max(\pw{t_0}_1,r^{-1}(\pt{t_0}))$.
\end{itemize} 
However, in multi-dimsnional case, since we changed the true values $\theta_{\star}$ from $\ptpa{t}$ to $\ptpa{t+1}$ after each iteration, definition of $R_5$ and $R_6$ changes and relation (a) in \eqref{eq:wbound_eq1} does not hold anymore, namely, 
$$D(\tpt{t+1}_1,\pw{t+1}_1;\ptpa{t+1})\not\subset D(\|\vpt{t}\|,\pw{t}_1;\ptpa{t}).$$    
Yet, we can have a quick remedy for this strategy. Note that since $\ptpa{t}\rightarrow \|\vtheta^*\|$, our adjusted reference curve $r_{adj}(w_1;\ptpa{t})$ also converges to $r_{adj}(w_1;\|\vtheta^*\|)$ uniformly for $w_1\in [w^*_1,1]$. Hence, we can find $\delta'>0$, $t'>0$ such that we can perturb every $r_{adj}(w_1;\ptpa{t})$
 for $t>t'$ such that we have $\tilde{r}_{adj}(w_1;\ptpa{t})$ satisfies C.1 and C.2 for true value $\theta_{\star}=\ptpa{t}$ for all $t>t'$ with
 $$\tilde{r}_{adj}(w_1;\theta_{\star})\= r_{adj}(w_1;\ptpa{t'}),\quad \forall w_1\in [1-\delta',1], \theta_{\star}\in [\ptpa{t'},\|\vtheta^*\|],$$
 and
 $$\tilde{r}_{adj}(w_1;\theta_{\star})\= r(w_1;\theta_{\star}), \quad \forall w_1\leq w^*_1, \theta_{\star}\in [\ptpa{t'},\|\vtheta^*\|].$$
 Hence, the region $R_5$ and $R_6$ are invariant for $\theta_{\star}\in [\ptpa{t'},\|\vtheta^*\|]$, and therefore with the same arguments made in the proof of Lemma \ref{lem:converge}, within finite iterations $t''$, we have 
 $$\|\vpt{t''}\|>\ptpa{t'}(w^*_1-w^*_2),$$ 
 in other words, $(\|\vpt{t''}\|,\pw{t''}_1)$ lies in $R_1\bigcup R_2\bigcup R_3\bigcup R_4$ for any true value $\theta_{\star} \in [\ptpa{t'},\|\vtheta\|^*]$. Once the point $(\|\vpt{t''}\|,\pw{t''}_1)$ lies in the region $R_1\bigcup R_2\bigcup R_3\bigcup R_4$, we can bound every $(\|\vpt{t+1}\|,\pw{t+1}_1)$ for all $t\geq t''$ by 
 \begin{eqnarray}
 D\left(\min\left(\tilde{r}_{adj}(1-\delta'),\|\vpt{t}\|\right),\min\left(c_3,r^{-1}(c_2;\ptpa{t})\right);\|\vtheta^*\|\right)\bigcup D\left(c_2,\max(\pw{t}_1,1-\delta');\ptpa{t}\right),\label{eq:qset}
 \end{eqnarray}
 due to the fact that $(\tpt{t+1}_1,\pw{t+1}_1)\in D(\|\vpt{t}\|,\pw{t}_1;\ptpa{t})$ and $\|\vpt{t+1}\|\leq c_2$. Denote the set defined in \eqref{eq:qset} as $Q(\|\vpt{t}\|,\pw{t}_1)$. Then, we can check that for any $(\theta,w_1)\in Q(\|\vpt{t}\|,\pw{t}_1)$, we have $Q(\theta,w_1)\subseteq Q(\|\vpt{t}\|,\pw{t}_1)$. Therefore, we have $Q(\|\vpt{t+1}\|,\pw{t+1}_1)\subseteq Q(\|\vpt{t}\|,\pw{t})$. Hence, by a chain of arguments starting from $t''$, we have
 $$(\|\vpt{t+1}\|,\pw{t+1}_1) \in Q((\|\vpt{t''}\|,\pw{t''}_1)).$$
 Hence, we have 
 $$\pw{t}_1\leq \max\left(\tilde{r}_{adj}^{-1}(\|\vpt{t''}\|;\|\vtheta^*\|),1-\delta',\pw{t''}_1\right)\<1, \quad \forall t\geq t''.$$



\subsection{Geometric convergence} \label{sec:attraction}

Since we have shown that $(\vpt{t},\pw{t})$ converges to $(\vtheta^*,w_1^*)$, we just need to show an attraction basin around $(\vtheta^*,w_1^*)$, and therefore, combining both, we know after a finite iteration $T$, we have geometric convergence. To show an attraction basin, let us consider the following two terms $\|\vpt{t+1}-\vtheta^*\|$ and $|\pw{t+1}_1-w_1^*|$. Note that, at iteration $t$, let us choose the coordinate such that $\vpt{t}=(\|\vpt{t}\|,0,\ldots,0)$ and $\vtheta^*=(\ptpa{t},\ptpe{t},0,\ldots,0)$, then by \eqref{eq:mo_eq1} and \eqref{eq:s}, we have
\begin{eqnarray}
	\|\vpt{t+1}-\vtheta^*\|^2
	&=&
	|\tpt{t+1}_1-\ptpa{t}|^2+|\tpt{t+1}_2-\ptpe{t}|^2
		\n\\
	&=&
	|G_{\theta}(\|\vpt{t}\|,\pw{t}_1;\ptpa{t},w^*_1)-\ptpa{t}|^2+|\ptpe{t}|^2(1-s(\|\vpt{t}\|,\pw{t}_1;\ptpa{t},w^*_1))^2
	,	\n\\
	|\pw{t+1}_1-w_1^*|
	&=&
	|G_{w}(\|\vpt{t}\|,\pw{t}_1;\ptpa{t},w^*_1)-w_1^*|
	.	\label{eq:basin_eq1}
\end{eqnarray}
Hence, we just need to show that for all $\theta^*_{\parallel}>0$ and $w^*_1\in (0,1)$, the eigenvalues of the Jacobian matrix of the following mapping:
\begin{eqnarray}
	(\theta,w_1)\mapsto (G_{\theta}(\theta,w_1;\theta^*_{\parallel},w^*_1),G_{w}(\theta,w_1;\theta^*_{\parallel},w^*_1)) \label{eq:mapping}
\end{eqnarray}
are in $[0,1)$ at $(\theta,w_1)=(\theta^*_{\parallel},w^*_1)$. Then, note that 
\[
	G_{\theta}(\theta^*_{\parallel},w^*_1;\theta^*_{\parallel},w^*_1)
	\=
	\theta^*_{\parallel}
	\kand
	G_{w}(\theta^*_{\parallel},w^*_1;\theta^*_{\parallel},w^*_1)
	\=
	w^*_1.
\]
Hence, by continuity of the Jacobian of the functions, there exists $\epsilon>0$ and $\rho<1$ such that as long as $\theta,\theta^*_{\parallel} \in [\|\vtheta^*\|-\epsilon,\|\vtheta^*\|+\epsilon]$ and $w_1\in [w_1^*-\epsilon,w_1^*+\epsilon]$, we have
\[
	(G_{\theta}(\theta,w_1;\theta^*_{\parallel},w^*_1)-\theta^*_{\parallel})^2+(G_{w}(\theta,w_1;\theta^*_{\parallel},w^*_1)-w^*_1)^2
	\kleq
	\rho\left((\theta-\theta^*_{\parallel})^2+(w_1-w^*_1)^2\right)
	.
\]
Further, by \eqref{eq:initial_eq5}, we know function $s(\theta,w_1;\theta^*_{\parallel},w^*_1)$ is positive on $\theta,\theta^*_{\parallel} \in [\|\vtheta^*\|-\epsilon,\|\vtheta^*\|+\epsilon]$ and $w_1\in [w_1^*-\epsilon,w_1^*+\epsilon]$. Hence, there exists constant $\rho'$ such that 
\[
	(1-s(\theta,w_1;\theta^*_{\parallel},w^*_1))^2
	\kleq
	\rho'
	,	\quad \forall  \theta,\theta^*_{\parallel} \in [\|\vtheta^*\|-\epsilon,\|\vtheta^*\|+\epsilon], w_1\in [w_1^*-\epsilon,w_1^*+\epsilon]
	.
\]
Hence, plug in \eqref{eq:basin_eq1}, we have if $\|\vpt{t}\|,\ptpa{t} \in [\|\vtheta^*\|-\epsilon,\|\vtheta^*\|+\epsilon]$ and $\pw{t}_1\in [w_1^*-\epsilon,w_1^*+\epsilon]$, then
\begin{eqnarray}
	\|\vpt{t+1}-\vtheta^*\|^2+|\pw{t+1}_1-w_1^*|^2
	&\leq&
	\rho\left((\|\vpt{t}\|-\ptpa{t})^2+(\pw{t}_1-w^*_1)^2\right)+\rho'|\ptpe{t}|^2
		\n\\
	&\leq&
	\max(\rho,\rho')\left(\|\vpt{t}-\vtheta^*\|^2+(\pw{t}_1-w^*_1)^2\right)
	.	\n
\end{eqnarray}
Hence, by triangle inequality, we know once $\|\vpt{t}-\vtheta^*\|\leq \epsilon$ and $|\pw{t}_1-w_1^*|\leq \epsilon$, we have $(\vpt{t},\pw{t}_1)$ geometrically converges towards $(\vtheta^*,w_1^*)$. Further, the first iteration to reach the attraction basin is guaranteed by the geometric convergence of the angle $\pb{t}$ and geometric convergence of the area function $m(\theta,w)$ on $S'$ defined in \eqref{eq:S'} for $\delta_0=\epsilon/4$.  

Next, we will show that for all $\theta^*_{\parallel}>0$ and $w^*_1\in (0,1)$, the eigenvalues of the Jacobian matrix of the mapping defined in \eqref{eq:mapping} at $(\theta,w_1)=(\theta^*_{\parallel},w^*_1)$ are in $[0,1)$. Note that this Jacobian matrix at $(\theta,w_1)=(\theta^*_{\parallel},w^*_1)$ is the following:
\begin{eqnarray}
	J
	&=&
	\left[\begin{aligned}
	&\underbrace{\int \frac{4w^*_1w^*_2y^2}{w^*_1e^{y\theta^*_{\parallel}}+w^*_2e^{-y\theta^*_{\parallel}}}\phi(y)e^{-\frac{(\theta^*_{\parallel})^2}{2}}\dif y}_{J_{11}} 
	&&\underbrace{\int \frac{2y}{w^*_1e^{y\theta^*_{\parallel}}+w^*_2e^{-y\theta^*_{\parallel}}}\phi(y)e^{-\frac{(\theta^*_{\parallel})^2}{2}}\dif y}_{J_{12}}
		\\
	&\underbrace{\int \frac{2w^*_1w^*_2y}{w^*_1e^{y\theta^*_{\parallel}}+w^*_2e^{-y\theta^*_{\parallel}}}\phi(y)e^{-\frac{(\theta^*_{\parallel})^2}{2}}\dif y}_{J_{21}} 
	&&\underbrace{\int \frac{1}{w^*_1e^{y\theta^*_{\parallel}}+w^*_2e^{-y\theta^*_{\parallel}}}\phi(y)e^{-\frac{(\theta^*_{\parallel})^2}{2}}\dif y}_{J_{22}}
	\end{aligned}\right]
	.	\n
\end{eqnarray}
Then the two eigenvalues of $J$ should be the two solutions of the following equation:
\[
	q(\lambda)
	\:=\
	\lambda^2-\lambda(J_{11}+J_{22})+J_{11}J_{22}-J_{12}J_{21}
	\=
	0
	.
\]
Note that, by Cauchy inequality, we know $\text{det}(J)=J_{11}J_{22}-J_{12}J_{21}\geq 0$ and therefore $q(0)\geq 0$. Also note that 
\[
	q(J_{22})
	\=
	-J_{22}^2-J_{12}J_{21}
	\kleq
	0,
\] 
and
\begin{eqnarray}
	0\ < \ J_{22}
	&=&\int_{y\geq 0} \frac{e^{y\theta^*_{\parallel}}+e^{-y\theta^*_{\parallel}}}{w^*_1w^*_2(e^{y\theta^*_{\parallel}}-e^{-y\theta^*_{\parallel}})^2+1}\phi(y)e^{-\frac{(\theta^*_{\parallel})^2}{2}}\dif y
		\n\\
	&=&
	\int_{y\geq 0} (e^{y\theta^*_{\parallel}}+e^{-y\theta^*_{\parallel}})\phi(y)e^{-\frac{(\theta^*_{\parallel})^2}{2}}\dif y
		\n\\
	&&
	\quad \quad \quad\quad \quad \quad 
	-\int_{y\geq 0} \frac{w^*_1w^*_2(e^{y\theta^*_{\parallel}}+e^{-y\theta^*_{\parallel}})(e^{y\theta^*_{\parallel}}-e^{-y\theta^*_{\parallel}})^2}{w^*_1w^*_2(e^{y\theta^*_{\parallel}}-e^{-y\theta^*_{\parallel}})^2+1}\phi(y)e^{-\frac{(\theta^*_{\parallel})^2}{2}}\dif y
		\n\\
	&=&
	1-\int_{y\geq 0} \frac{w^*_1w^*_2(e^{y\theta^*_{\parallel}}+e^{-y\theta^*_{\parallel}})(e^{y\theta^*_{\parallel}}-e^{-y\theta^*_{\parallel}})^2}{w^*_1w^*_2(e^{y\theta^*_{\parallel}}-e^{-y\theta^*_{\parallel}})^2+1}\phi(y)e^{-\frac{(\theta^*_{\parallel})^2}{2}}\dif y
		\n\\
	&\leq&
	1
	.	\label{eq:basin_eq2}
\end{eqnarray}
Hence, we just need to show $q(1)>0$, then the two solutions of $q(\lambda)=0$ should stay in $[0,1)$. Note that
\begin{eqnarray}
	J_{11}
	&=&
	\int_{y\geq 0} \frac{4w_1^*w_2^*(e^{y\theta^*_{\parallel}}+e^{-y\theta^*_{\parallel}})y^2}{w^*_1w^*_2(e^{y\theta^*_{\parallel}}-e^{-y\theta^*_{\parallel}})^2+1}\phi(y)e^{-\frac{(\theta^*_{\parallel})^2}{2}}\dif y
		\n\\
	&=&
	\int_{y\geq 0} \frac{4y^2}{e^{y\theta^*_{\parallel}}+e^{-y\theta^*_{\parallel}}}\phi(y)e^{-\frac{(\theta^*_{\parallel})^2}{2}}\dif y
		\n\\
	&&
	\quad \quad \quad\quad \quad \quad 
	-\int_{y\geq 0} \frac{4(w_1^*-w_2^*)^2y^2}{(e^{y\theta^*_{\parallel}}+e^{-y\theta^*_{\parallel}})(w^*_1w^*_2(e^{y\theta^*_{\parallel}}-e^{-y\theta^*_{\parallel}})^2+1)}\phi(y)e^{-\frac{(\theta^*_{\parallel})^2}{2}}\dif y
		\n\\
	&<&
	1-\int_{y\geq 0} \frac{4(w_1^*-w_2^*)^2y^2}{(e^{y\theta^*_{\parallel}}+e^{-y\theta^*_{\parallel}})(w^*_1w^*_2(e^{y\theta^*_{\parallel}}-e^{-y\theta^*_{\parallel}})^2+1)}\phi(y)e^{-\frac{(\theta^*_{\parallel})^2}{2}}\dif y
	,	\label{eq:basin_eq3}
\end{eqnarray} 
where the last inequality holds due to the fact that
\[
	\int_{y\geq 0} \frac{4y^2}{e^{y\theta^*_{\parallel}}+e^{-y\theta^*_{\parallel}}}\phi(y)e^{-\frac{(\theta^*_{\parallel})^2}{2}}\dif y
	\kleq
	\int_{y\geq 0} 2y^2\phi(y)e^{-\frac{(\theta^*_{\parallel})^2}{2}}\dif y
	\=
	e^{-\frac{(\theta^*_{\parallel})^2}{2}}
	.
\]
Combine \eqref{eq:basin_eq2} and \eqref{eq:basin_eq3}, we have
\begin{eqnarray}
	q(1)
	&=&
	(1-J_{11})(1-J_{22})-J_{12}J_{21}
		\n\\
	&>& \int_{y\geq 0} \frac{4(w_1^*-w_2^*)^2y^2}{(e^{y\theta^*_{\parallel}}+e^{-y\theta^*_{\parallel}})(w^*_1w^*_2(e^{y\theta^*_{\parallel}}-e^{-y\theta^*_{\parallel}})^2+1)}\phi(y)e^{-\frac{(\theta^*_{\parallel})^2}{2}}\dif y
		\n\\
	&& \times \int_{y\geq 0} \frac{w^*_1w^*_2(e^{y\theta^*_{\parallel}}+e^{-y\theta^*_{\parallel}})(e^{y\theta^*_{\parallel}}-e^{-y\theta^*_{\parallel}})^2}{w^*_1w^*_2(e^{y\theta^*_{\parallel}}-e^{-y\theta^*_{\parallel}})^2+1}\phi(y)e^{-\frac{(\theta^*_{\parallel})^2}{2}}\dif y
		\n\\
	&&-4w_1^*w_2^*(w_1^*-w_2^*)^2\int_{y\geq 0} \left(\frac{(e^{y\theta^*_{\parallel}}-e^{-y\theta^*_{\parallel}})y}{w^*_1w^*_2(e^{y\theta^*_{\parallel}}-e^{-y\theta^*_{\parallel}})^2+1}\phi(y)e^{-\frac{(\theta^*_{\parallel})^2}{2}}\dif y\right)^2
		\n\\
	&\geq&
	0
	,	\n
\end{eqnarray}
where the last inequality holds due to Cauchy inequality. Hence, we have $q(1)>0$ and this completes our proof for geometric convergence of the EM estimates.
 
 
\section{Proof of Theorem \ref{thm:landscape}}\label{sec:landscape}

The maximum log-likelihood objective for population-${\rm EM}_2$ is the following optimization problem:
\begin{eqnarray}
	\max_{\vtheta\in \bbR^d, w_1\in [0,1]} \bbE_{\vy\sim f^*} \log\left(w_1e^{-\frac{\|\vy-\vtheta\|^2}{2}}+w_2e^{-\frac{\|\vy+\vtheta\|^2}{2}}\right)
	.	\label{eq:land_eq1}
\end{eqnarray}
Due to the symmetric property of the landscape, without loss of generality, we assume $w_1^*>0.5$.
Note that the first order stationary points of above optimization problem should satisfy the following equation. 
\begin{eqnarray}
	\bbE_{\vy\sim f^*}\left[\frac{w_1e^{\dotp{\vy,\vtheta}}-w_2e^{-\dotp{\vy,\vtheta}}}{w_1e^{\dotp{\vy,\vtheta}}+w_2e^{-\dotp{\vy,\vtheta}}}\vy\right] - \vtheta
	&=&
	\v0
	,	\label{eq:land_eq3}\\
	\bbE_{\vy\sim f^*}\left[\frac{e^{\dotp{\vy,\vtheta}}-e^{-\dotp{\vy,\vtheta}}}{w_1e^{\dotp{\vy,\vtheta}}+w_2e^{-\dotp{\vy,\vtheta}}}\right] 
	&=&
	0
	.	\label{eq:land_eq4}
\end{eqnarray}
We first consider the two trivial cases when $w_1=1$ and $w_1=0$. Suppose $w_1=1$, then from \eqref{eq:land_eq3}, we have $\vtheta = (w_1^*-w_2^*)\vtheta^*$. Hence, plug it in \eqref{eq:land_eq4}, we have the following equation holds
\begin{eqnarray}
	\int\left(1-e^{-2(w_1^*-w_2^*)y\|\vtheta^*\|}\right)\left(w_1^*\phi(y-\|\vtheta^*\|)+w_2^*\phi(y+\|\vtheta^*\|)\right)\dif y
 	&=&
	0
	,	\n
\end{eqnarray}
which is equivalent to
\begin{eqnarray}
	1-w_1^* e^{-4w_2^*(w_1^*-w_2^*)\|\vtheta^*\|^2}-w_2^*e^{4w_1^*(w_1^*-w_2^*)\|\vtheta^*\|^2}
	&=&
	0
	.	\n
\end{eqnarray}
Taking the derivative with respect to $\|\vtheta^*\|$, it is straightforward to show that when $w_1^*>0.5$, the LHS is a strictly decreasing function of $\|\vtheta^*\|$ and achieves its maximum 0 at $\|\vtheta^*\|=0$. Hence, it contradicts the RHS of the equation and therefore \eqref{eq:land_eq3} and \eqref{eq:land_eq4} can not hold simultaneously for $w_1=1$. Hence, there is no first order stationary point for the case $w_1=1$ and similarly for $w_1=0$. 

Now we restrict $w_1\in (0,1)$. Then it is straightforward to show that every first order stationary point of the optimization in \eqref{eq:land_eq1} should be a fixed point for population-${\rm EM}_2$. From the proof of Theorem \ref{thm:main}, we know the two global maxima $(\vtheta^*,w_1)$ and $(-\vtheta^*,w_2)$ are the only fixed points of population-${\rm EM}_2$ in the following region:
\[
	\underbrace{\left\{(\vtheta,w_1)| w_1\in [0.5,1), \dotp{\vtheta,\vtheta^*}>0\right\}}_{\text{Area}_1} \bigcup \underbrace{\left\{(\vtheta,w_1)| w_1\in (0,0.5], \dotp{\vtheta,\vtheta^*}<0\right\}}_{\text{Area}_2}
\] 
Furthermore, for any fixed point lies in the hyperplane $\mathcal{H}: \dotp{\vtheta,\vtheta^*}=0$, it is clear that its corresponding $w_1$ should be $0.5$. Further, since $\dotp{\vtheta,\vtheta^*}=0$, from \eqref{eq:land_eq3}, it is clear that $\vtheta$ should satisfy the following equation
\begin{eqnarray}
	\int \frac{e^{y\|\vtheta\|}-e^{-y\|\vtheta\|}}{e^{y\|\vtheta\|}+e^{-y\|\vtheta\|}}y\phi(y)\dif y
	&=&
	\|\vtheta\|
	.	\n
\end{eqnarray}
Since the derivative with respect to $\|\vtheta\|$ of the LHS is in $(0,1)$ for $\|\vtheta\|>0$, it is clear that $\|\vtheta\|=0$ is the only solution for the equation and therefore, $(\vtheta,w_1)=(\v0,\frac{1}{2})$ is the only fixed point in the hyperplane $\mathcal{H}$. Furthermore, the Hessian of the log-likelihood in \eqref{eq:land_eq1} at $(\vtheta,w_1)=(\v0,\frac{1}{2})$ is the following matrix.
\begin{eqnarray}
	\left[\begin{aligned}
		\vtheta^*(\vtheta^*)^{\t}&&\quad	 2(w_1^*-w_2^*)\vtheta^*&&\\
		2(w_1^*-w_2^*)(\vtheta^*)^{\t}&& \quad 0&&
		\end{aligned}\right]
\end{eqnarray}
It is clear that it has a positive eigenvalue, a negative eigenvalue and therefore $(\v0,\frac{1}{2})$ is a saddle point. 

Finally, we will show there is no fixed point in the rest of the region in $\bbR^2\times [0,1]$, i.e.,
\[
	\underbrace{\left\{(\vtheta,w_1)| w_1\in (0,0.5), \dotp{\vtheta,\vtheta^*}>0\right\}}_{\text{Area}_3}\bigcup \underbrace{\left\{(\vtheta,w_1)| w_1\in (0.5,1), \dotp{\vtheta,\vtheta^*}<0\right\}}_{\text{Area}_4}
\] 
Due to the symmetric property, we will just prove the result for $\text{Area}_3$. Note that, by Lemma \ref{lem:shapeg} and the fact that 
\begin{eqnarray}
	g_{w}(\theta,0.5)
	\ \lessgtr \
	0.5
	,	\quad \forall \theta \lessgtr 0
	.
\end{eqnarray} 
We know for all $w_1\in (0,0.5)$, 
\begin{eqnarray}
	0
	&<&
	g_w(\|\vtheta\|,w_1; \theta_{\parallel},w_1^*)-w_1
		\n\\
	&=&
	w_1w_2\int \left[\frac{e^{y\|\vtheta\|}-e^{-y\|\vtheta\|}}{w_1e^{y\|\vtheta\|}+w_2e^{-y\|\vtheta\|}}\right]\left(w^*_1\phi(y-\theta_{\parallel})+w^*_2\phi(y+\theta_{\parallel})\right)\dif y
		\n\\
	&=&
	w_1w_2\cdot \bbE_{\vy\sim f^*}\left[\frac{e^{\dotp{\vy,\vtheta}}-e^{-\dotp{\vy,\vtheta}}}{w_1e^{\dotp{\vy,\vtheta}}+w_2e^{-\dotp{\vy,\vtheta}}}\right]
	,	\n 
\end{eqnarray}
where $\theta_{\parallel}=\dotp{\vtheta^*,\vtheta}/\|\vtheta\|$. Hence, there is no solution for \eqref{eq:land_eq4} in $\text{Area}_3$. This completes the proof of this theorem.

\section{Proof of Theorem \ref{thm:fsa}}\label{sec:fsa}
 
Let $(\hvpt{t},\hpw{t}_1)$ denote the finite sample estimate. To show the convergence of the finite sample estimate, we want to argue that its behavior is close to the corresponding convergence behavior of the population estimate. Hence, let us first prove the following uniform concentration bounds that for any fixed constant $c>0$, with probability at least $1-\delta$, we have
\begin{eqnarray}
	\Delta_w:=\sup_{\|\vtheta\|\in [0,c],w_1\in[0,1]}\left|\frac{1}{n}\sum_{i=1}^n\left[\frac{w_1e^{\dotp{\vy_i,\vtheta}}}{w_1e^{\dotp{\vy_i,\vtheta}}+w_2e^{-\dotp{\vy_i,\vtheta}}}\right]-\bbE_{\vy\sim f^*}\left[\frac{w_1e^{\dotp{\vy,\vtheta}}}{w_1e^{\dotp{\vy,\vtheta}}+w_2e^{-\dotp{\vy,\vtheta}}}\right]\right|
	&&
		\n\\
	\kleq
	O\left((\|\vtheta^*\|+1)\sqrt{\frac{d+\ln(2/\delta)}{n}}\right)
	&&	\label{eq:finite_eq1}\\
	\Delta_{\theta}:=\sup_{\|\vtheta\|\in [0,c],w_1\in[0,1]}\left\|\frac{1}{n}\sum_{i=1}^n\left[\frac{w_1e^{\dotp{\vy_i,\vtheta}}-w_2e^{-\dotp{\vy_i,\vtheta}}}{w_1e^{\dotp{\vy_i,\vtheta}}+w_2e^{-\dotp{\vy_i,\vtheta}}}\vy_i\right]-\bbE_{\vy\sim f^*}\left[\frac{w_1e^{\dotp{\vy,\vtheta}}-w_2e^{-\dotp{\vy,\vtheta}}}{w_1e^{\dotp{\vy,\vtheta}}+w_2e^{-\dotp{\vy,\vtheta}}}\vy\right]\right\|
	&&
		\n\\
	\kleq
	O\left((\|\vtheta^*\|+1)\sqrt{\frac{d+\ln(2/\delta)}{n}}\right)
	.
	&&	\label{eq:finite_eq2}
\end{eqnarray}  
To show \eqref{eq:finite_eq1}, by Jensen's inequality, we have
\begin{eqnarray}
	\bbE e^{\lambda \Delta_w}
	&\leq&
	\bbE_{y,y'}\text{exp}\left(\lambda\sup_{\|\vtheta\|\in [0,c],w_1\in[0,1]}\left|\frac{1}{n}\sum_{i=1}^n\left(\frac{w_1e^{\dotp{\vy_i,\vtheta}}}{w_1e^{\dotp{\vy_i,\vtheta}}+w_2e^{-\dotp{\vy_i,\vtheta}}}-\frac{w_1e^{\dotp{\vy'_i,\vtheta}}}{w_1e^{\dotp{\vy'_i,\vtheta}}+w_2e^{-\dotp{\vy'_i,\vtheta}}}\right)\right|\right).\n
\end{eqnarray} 
Then, we introduce i.i.d.~Rademacher variables $\xi_i$ and obtain that
\begin{eqnarray}
	\bbE e^{\lambda \Delta_w}
	&\leq&
	\bbE_{y,\xi}\text{exp}\left(2\lambda\sup_{\|\vtheta\|\in [0,c],w_1\in[0,1]}\left|\frac{1}{n}\sum_{i=1}^n\xi_i\left(\frac{w_1e^{\dotp{\vy_i,\vtheta}}}{w_1e^{\dotp{\vy_i,\vtheta}}+w_2e^{-\dotp{\vy_i,\vtheta}}}-w_1\right)\right|\right)
	.	\n
\end{eqnarray} 
Now apply the following lemma from \cite{VKoltchinskii2011tailbound}
\begin{lemma}\label{lem:symmetrize}
Let $\mathcal{H}\in\bbR^n$ and let $\psi_i: \bbR\mapsto \bbR, i=1, \cdots, n$ be functions such that $\psi_i(0)=0$ and
\[
	|\psi_i(u)-\psi_i(v)|
	\kleq
	|u-v|
	\in \bbR.
\]
For all convex nondecreasing functions $\Psi:\bbR_+\mapsto \bbR_+$,
\[
	\bbE\Psi(\frac{1}{2}\sup_{\vh\in \mathcal{H}}|\sum_{i=1}^n\psi_i(h_i)\xi_i|)
	\kleq
	\bbE\Psi(\sup_{\vh\in \mathcal{H}}|\sum_{i=1}^nh_i\xi_i|)
	, 
\]
where $\xi_i$ are i.i.d.~Rademacher random variables.
\end{lemma}
We have
\begin{eqnarray}
	\bbE e^{\lambda \Delta_w}
	&\leq&
	\bbE_{y,\xi}\text{exp}\left(2\lambda\sup_{\|\vtheta\|\in [0,c],w_1\in[0,1]}\left|\frac{1}{n}\sum_{i=1}^n\xi_i\dotp{\vy_i,\vtheta}\right|\right)
		\n\\
	&\leq&
	\bbE_{y,\xi}\text{exp}\left(2\lambda c\left\|\frac{1}{n}\sum_{i=1}^n\xi_i\vy_i\right\|\right)
		\n\\
	&=&
	\bbE_{\tilde{y}}\text{exp}\left(2\lambda c\left\|\frac{1}{n}\sum_{i=1}^n\tilde{\vy}_i\right\|\right)
	,	\n
\end{eqnarray}
where $\tilde{\vy}_i$ are i.i.d.~random variables following this symmetric distribution: $\frac{1}{2}\mathcal{N}(-\vtheta^*,\vI)+\frac{1}{2}\mathcal{N}(\vtheta^*,\vI)$. Then apply a typical argument of $1/2$-covering net over the $d$-dimensional unit sphere, it is straight forward to show that we have
 \begin{eqnarray}
	\bbE e^{\lambda \Delta_w}
	&\leq&
	\text{exp}\left(8\lambda^2 c^2\frac{\|\vtheta^*\|^2+1}{n}+2d\right)
	.	\n
\end{eqnarray}
Apply Markov inequality and choose $\lambda$ properly, we have \eqref{eq:finite_eq1} holds. To prove \eqref{eq:finite_eq2}, we follow the proof of corollary 2 in B.2 in \cite{balakrishnan2014statistical}. Let 
\[
	\Delta_{\theta}^{\vu}
	\=
	\sup_{\|\vtheta\|\in [0,c],w_1\in[0,1]}\frac{1}{n}\sum_{i=1}^n\left[\frac{w_1e^{\dotp{\vy_i,\vtheta}}-w_2e^{-\dotp{\vy_i,\vtheta}}}{w_1e^{\dotp{\vy_i,\vtheta}}+w_2e^{-\dotp{\vy_i,\vtheta}}}\right]\dotp{\vy_i,\vu}-\bbE_{\vy\sim f^*}\left[\frac{w_1e^{\dotp{\vy,\vtheta}}-w_2e^{-\dotp{\vy,\vtheta}}}{w_1e^{\dotp{\vy,\vtheta}}+w_2e^{-\dotp{\vy,\vtheta}}}\right]\dotp{\vy,\vu}
	.
\]
Then, we have
\begin{eqnarray}
	\bbE e^{\lambda \Delta_{\theta}}
	&=&
	\bbE_y e^{\lambda \sup_{\|\vu\|=1}\Delta_{\theta}^{\vu}}
	\kleq
	\bbE_y e^{2\lambda\max_{j\in[M]}\Delta_{\theta}^{\vu_j}}
	\kleq
	\sum_{j=1}^M\bbE_y e^{2\lambda \Delta_{\theta}^{\vu_j}}
		\n\\
	&\leq&
	\sum_{j=1}^M\bbE_{y, \xi}\text{exp}\left(4\lambda \sup_{\|\vtheta\|\in [0,c],w_1\in[0,1]}\frac{1}{n}\sum_{i=1}^n\xi_i
	\left[\frac{w_1e^{\dotp{\vy_i,\vtheta}}-w_2e^{-\dotp{\vy_i,\vtheta}}}{w_1e^{\dotp{\vy_i,\vtheta}}+w_2e^{-\dotp{\vy_i,\vtheta}}}-(w_1-w_2)\right]\dotp{\vy_i, \vu_{j}}\right)
	, \n
\end{eqnarray}
where $\{\vu_j\}_{j=1}^M$ is the $\frac{1}{2}$-covering net over the $d$ dimensional unit sphere and $\xi_i$ are i.i.d.~Rademacher random variables and the last inequality holds for standard symmetrization result for empirical process. Apply Lemma \ref{lem:symmetrize} again, we have
\begin{eqnarray}
	\bbE e^{\lambda \Delta_{\theta}}
	&\leq&
	\sum_{j=1}^M\bbE_{y, \xi}\text{exp}\left(4\lambda \sup_{\|\vtheta\|\in [0,c],w_1\in[0,1]}\frac{1}{n}\sum_{i=1}^n\xi_i
	\dotp{\vy_i,\vtheta}\dotp{\vy_i, \vu_{j}}\right)
		\n\\
	&\leq&
	e^{2d}\cdot \bbE_{y, \xi}\text{exp}\left(4\lambda  c\left\|\frac{1}{n}\sum_{i=1}^n\xi_i \vy_i\vy_i^{\t}
	\right\|_{\text{op}}\right)
	,	\n
\end{eqnarray}
where $\| \cdot \|_{\text{op}}$ is the $\ell_2$-operator norm of a matrix (the maximum singular value). Follow the result in B.2 in \cite{balakrishnan2014statistical}, we have
\begin{eqnarray}
	\bbE_{y, \xi}\text{exp}\left(4\lambda  c\left\|\frac{1}{n}\sum_{i=1}^n\xi_i \vy_i\vy_i^{\t}
	\right\|_{\text{op}}\right)
	&\leq&
	\sum_{j=1}^M \bbE_{y, \xi}\text{exp}\left(8\lambda  c \frac{1}{n}\sum_{i=1}^n\xi_i \dotp{\vy_i,\vu_j}^2\right)
		\n\\
	&=&
	\sum_{j=1}^M \bbE_{y, \xi,\xi'}\text{exp}\left(8\lambda  c \frac{1}{n}\sum_{i=1}^n\xi_i \dotp{\xi_i'\vy_i,\vu_j}^2\right)
		\n\\
	&=&
	\sum_{j=1}^M \bbE_{\tilde{y}, \xi}\text{exp}\left(8\lambda  c \frac{1}{n}\sum_{i=1}^n\xi_i \dotp{\tilde{\vy}_i,\vu_j}^2\right)
	,	\n
\end{eqnarray} 
where $\xi_i'$ are independent copies of Rademacher random variables. Hence, from \cite{balakrishnan2014statistical}, we have
\begin{eqnarray}
	\sum_{j=1}^M \bbE_{\tilde{y}, \xi}\text{exp}\left(8\lambda  c \frac{1}{n}\sum_{i=1}^n\xi_i \dotp{\tilde{\vy}_i,\vu_j}^2\right)
	&\leq&
	e^{\frac{32\lambda^2c^2(\|\vtheta^*\|^2+1)}{n}+2d}
	. 	\n
\end{eqnarray}
Hence, combine all, we have
\[
	\bbE e^{\lambda \Delta_{\theta}}
	\kleq
	e^{\frac{32\lambda^2c^2(\|\vtheta^*\|^2+1)}{n}+4d}
	.
\]
Apply Markov inequality and choose $\lambda$ properly, we have \eqref{eq:finite_eq2} holds.

Next, by choosing $c=\max(\|\hvpt{0}\|,2(1+\|\vtheta^*\|))$, it is straight forward to apply induction with Lemma \ref{lem:bound} to show that for sufficiently large $n$, with probability at least $1-\delta$,
\begin{eqnarray}
	\|\hvpt{t}\|
	&\leq&
	c
	, \quad \forall t\geq 0
	.	\n
\end{eqnarray}
Then, since the update functions are Lipchitz with constant at most $O(1+\|\vtheta^*\|)$, it is straight forward to show the following via induction that for any finite $t$,
\begin{eqnarray}
	\|\hvpt{t}-\vpt{t}\|^2+|\hpw{t}_1-\pw{t}_1|^2
	&\leq&
	O\left((1+\|\vtheta^*\|)^{t+1}\sqrt{\frac{d+\ln(2/\delta)}{n}}\right)
	. 	\n
\end{eqnarray}
From Appendix~\ref{sec:attraction}, we know there exists an attraction basin around $(\vtheta^*,w_1^*)$. Suppose this attraction basin contains the $\delta_0$-neighborhood around $(\vtheta^*,w_1^*)$, i.e., we have for some $\rho<1$,
\begin{eqnarray}
	\|\vpt{t+1}-\vtheta^*\|^2+|\pw{t+1}_1-w_1^*|^2
	\kleq
	\rho\left(\|\vpt{t}-\vtheta^*\|^2+(\pw{t}_1-w^*_1)^2\right)
	, \quad \forall \|\vpt{t}-\vtheta^*\|^2+|\pw{t}_1-w_1^*|^2\leq \delta_0^2	\n
\end{eqnarray}
Hence, from the proof in Appendix~\ref{sec:app_main}, we know there exists a finite iteration $T$ such that 
\[
	\|\vpt{T}-\vtheta^*\|^2+|\pw{T}_1-w_1^*|^2
	\kleq
	\frac{\delta_0^2}{2}
	,
\]
and therefore, for large enough $n$, with probability at least $1-\delta$, we have the finite sample estimate lies in the attraction basin after $T$ iteration, i.e.,
\[
	\|\hvpt{T}-\vtheta^*\|^2+|\hpw{T}_1-w_1^*|^2
	\kleq
	\delta_0^2
	.
\]
Once the finite sample estimate lies in the attraction basin, we follow the proof in \cite{balakrishnan2014statistical} and it is straight forward to show that for all $t\geq T$, we have
\[
	\|\hvpt{t}-\vtheta^*\|^2+|\hpw{t}_1-w_1^*|^2
	\kleq 
	\rho^{t-T}\left(\|\hvpt{T}-\vtheta^*\|^2+|\hpw{T}_1-w_1^*|^2\right)+O\left((\|\vtheta^*\|+1)\sqrt{\frac{d+\ln(2/\delta)}{n}}\right)
	.
\]
This completes our analysis for the convergence of the finite sample estimate.

\section{Proof of Auxiliary Lemmas}
\subsection{Proof of Lemma \ref{lem:Heq}}\label{sec:proofHeq}
In this proof, we have $w_1=w^*_1$. To prove \eqref{eq:Heq_eq1}, we just need to show
\begin{eqnarray}
 \frac{\partial h(\theta,w_1)}{\partial w_1}\left\{\begin{aligned}
			& >0,
			& &w_1>0.5 \\
			& <0,
			& &w_1<0.5
		\end{aligned}\right. \quad \forall \theta<\theta^*.\label{eq:Heq_eq3}
\end{eqnarray}
To prove this, we divide it into two cases (i) $\theta\leq 0$ and (ii) $\theta \in (0,\theta^*)$. To prove (i), by the definition of $h(\theta,w_1)$ in \eqref{eq:H} (with $w_2=1-w_1$), we have 
\begin{eqnarray}
\frac{\partial h(\theta,w_1)}{\partial w_1}&=&\underbrace{\int \frac{w_1e^{y\theta}-w_2e^{-y\theta}}{w_1e^{y\theta}+w_2e^{-y\theta}}y(\phi(y-\theta^*)-\phi(y+\theta^*))\dif y}_{\text{part}~ 1}\n\\
&&+2\underbrace{\int\frac{w_1e^{y\theta^*}+w_2e^{-y\theta^*}}{(w_1e^{y\theta}+w_2e^{-y\theta})^2}y\phi(y)e^{-\frac{(\theta^*)^2}{2}}\dif y}_{\text{part}~2}.\n
\end{eqnarray}
For part 1, we have
\begin{eqnarray}
\text{part}~1&=&\int_{y\geq 0}\left\{\frac{w_1e^{y\theta}-w_2e^{-y\theta}}{w_1e^{y\theta}+w_2e^{-y\theta}}+\frac{w_1e^{-y\theta}-w_2e^{y\theta}}{w_1e^{-y\theta}+w_2e^{y\theta}}\right\}y(\phi(y-\theta^*)-\phi(y+\theta^*))\dif y\n\\
&=&2\int_{y\geq 0}\frac{w_1^2-w_2^2}{w_1^2+w_2^2+w_1w_2(e^{-y\theta}+e^{y\theta})}y(\phi(y-\theta^*)-\phi(y+\theta^*))\dif y\n
\end{eqnarray}
Hence, we have
\begin{eqnarray}
\text{part}~ 1 \left\{\begin{aligned}
			& > 0,
			& &w_1>0.5 \\
			& < 0,
			& &w_1<0.5 \\
		\end{aligned}\right..\label{eq:Heq_eq4}
\end{eqnarray}
For part 2, we have
\begin{eqnarray}
\text{part}~2&=&\int_{y\geq 0}\left\{\frac{w_1e^{y\theta^*}+w_2e^{-y\theta^*}}{(w_1e^{y\theta}+w_2e^{-y\theta})^2}-\frac{w_1e^{-y\theta^*}+w_2e^{y\theta^*}}{(w_1e^{-y\theta}+w_2e^{y\theta})^2}\right\}y\phi(y)e^{-\frac{(\theta^*)^2}{2}}\dif y\n\\
&=&(w_1-w_2)\int_{y\geq 0} \left\{\frac{(w_1^2+w_2^2+w_1w_2)(e^{y(\theta^*-2\theta)}-e^{y(2\theta-\theta^*)})+2w_1w_2(e^{y\theta^*}-e^{-y\theta^*})}{(w_1e^{y\theta}+w_2e^{-y\theta})^2(w_1e^{-y\theta}+w_2e^{y\theta})^2}\right.\n\\
&&+\left.\frac{w_1w_2(e^{-y(\theta^*+2\theta)}-e^{y(\theta^*+2\theta)})}{(w_1e^{y\theta}+w_2e^{-y\theta})^2(w_1e^{-y\theta}+w_2e^{y\theta})^2}\right\}y\phi(y)e^{-\frac{(\theta^*)^2}{2}}\dif y.\n
\end{eqnarray}
Since $\theta\leq 0$, we have
\begin{eqnarray}
e^{y(\theta^*-2\theta)}-e^{y(2\theta-\theta^*)}&\geq& \max\left\{\left|e^{y\theta^*}-e^{-y\theta^*}\right|,\left|e^{y(\theta^*+2\theta)}-e^{-y(\theta^*+2\theta)}\right|\right\}.\n
\end{eqnarray}
Hence, we have
\begin{eqnarray}
\frac{\text{part}~ 2}{w_1-w_2}&\geq& \int_{y\geq 0} \frac{(w_1-w_2)^2(e^{y(\theta^*-2\theta)}-e^{y(2\theta-\theta^*)})}{(w_1e^{y\theta}+w_2e^{-y\theta})^2(w_1e^{-y\theta}+w_2e^{y\theta})^2}y\phi(y)e^{-\frac{(\theta^*)^2}{2}}\dif y \kgeq 0.\n
\end{eqnarray}
Therefore, we have
\begin{eqnarray}
\text{part}~ 2\left\{\begin{aligned}
			& \geq 0,
			& &w_1>0.5 \\
			& \leq 0,
			& &w_1<0.5 \\
		\end{aligned}\right..\label{eq:Heq_eq5}
\end{eqnarray}
Combine \eqref{eq:Heq_eq4} and \eqref{eq:Heq_eq5}, we have \eqref{eq:Heq_eq3} holds for case (i). To prove case (ii), we use a different strategy. First note that $h(\theta^*,w)\equiv \theta^*$, hence, 
\begin{eqnarray}
\frac{\partial h(\theta,w)}{\partial w}\Big|_{\theta=\theta^*}\= 0.\label{eq:Heq_eq6}
\end{eqnarray}
Therefore, to prove \eqref{eq:Heq_eq3} for case (ii), we just need to show 
\begin{eqnarray}
\frac{\partial^2 h(\theta,w_1)}{\partial \theta \partial w_1}\left\{\begin{aligned}
			& <0,
			& &w_1>0.5 \\
			& >0,
			& &w_1<0.5
		\end{aligned}\right.\quad \forall \theta\in (0,\theta^*).\label{eq:Heq_eq7}
\end{eqnarray}
By the definition of $h(\theta,w_1)$ in \eqref{eq:H} (with $w_2=1-w_1$), we have
\begin{eqnarray}
\frac{1}{4}\frac{\partial^2 h(\theta,w_1)}{\partial \theta\partial w_1}&=& \underbrace{2w_1w_2\int \frac{e^{y(\theta^*-\theta)}-e^{y(\theta-\theta^*)}}{(w_1e^{y\theta}+w_2e^{-y\theta})^3}y^2\phi(y)e^{-\frac{(\theta^*)^2}{2}}\dif y}_{\text{part}~3}\n\\
&&+
\underbrace{\int \left(-\frac{w_1^2e^{y\theta^*}}{(w_1e^{y\theta}+w_2e^{-y\theta})^2}+\frac{w_2^2e^{-y\theta^*}}{(w_1e^{y\theta}+w_2e^{-y\theta})^2}\right)y^2\phi(y)e^{-\frac{(\theta^*)^2}{2}}\dif y}_{\text{part}~4}\n
\end{eqnarray}
For part 3, we have
\begin{eqnarray}
\text{part}~ 3&=&2w_1w_2\int_{y\geq 0}\frac{(w_1-w_2)(e^{y(\theta^*-\theta)}-e^{y(\theta-\theta^*)})(e^{-y\theta}-e^{y\theta})(A^2+B^2-AB)}{(w_1e^{y\theta}+w_2e^{-y\theta})^3(w_1e^{-y\theta}+w_2e^{y\theta})^3}y^2\phi(y)e^{-\frac{(\theta^*)^2}{2}}\dif y,\n
\end{eqnarray}
where $A=w_1e^{y\theta}+w_2e^{-y\theta}$ and $B=w_1e^{-y\theta}+w_2e^{y\theta}$. Hence, since $\theta\in (0,\theta^*)$, we have
\begin{eqnarray}
\text{part}~ 3 \left\{\begin{aligned}
			& < 0,
			& &w_1>0.5 \\
			& > 0,
			& &w_1<0.5 \\
		\end{aligned}\right..\label{eq:Heq_eq8}
\end{eqnarray}
For part 4, we have
\begin{eqnarray}
\text{part}~ 4&=&-\int_{y\geq 0} (w_1-w_2) \frac{(w_1^2+w_2^2)(e^{(2\theta-\theta^*)y}+e^{-(2\theta-\theta^*)y})+2w_1w_2(e^{y\theta^*}+e^{-y\theta^*})}{(w_1e^{y\theta}+w_2e^{-y\theta})^2(w_1e^{-y\theta}+w_2e^{y\theta})^2}y^2\phi(y)e^{-\frac{(\theta^*)^2}{2}}\dif y.\n
\end{eqnarray}
Hence, we have
\begin{eqnarray}
\text{part}~ 4 \left\{\begin{aligned}
			& < 0,
			& &w_1>0.5 \\
			& > 0,
			& &w_1<0.5 \\
		\end{aligned}\right..\label{eq:Heq_eq9}
\end{eqnarray}
Combine \eqref{eq:Heq_eq8} and \eqref{eq:Heq_eq9}, we have \eqref{eq:Heq_eq7} holds and therefore \eqref{eq:Heq_eq3} holds for case (ii). This completes the proof for \eqref{eq:Heq_eq1}. To prove \eqref{eq:Heq_eq2}, note that
\begin{eqnarray}
0\ \leq \ \frac{\partial H(\theta,w_1)}{\partial \theta}&=& \int\frac{4w_1w_2}{(w_1e^{y\theta}+w_2e^{-y\theta})^2}y^2(w_1\phi(y-\theta^*)+w_2\phi(y+\theta^*))\dif y\n\\
&=&\underbrace{\int_{y\geq 0} \frac{4w_1w_2}{(w_1e^{y\theta}+w_2e^{-y\theta})^2}y^2(w_1\phi(y-\theta^*)+w_2\phi(y+\theta^*))\dif y}_{\text{part}~5}\n\\
&&+\underbrace{\int_{y\geq 0}\frac{4w_1w_2}{(w_2e^{y\theta}+w_1e^{-y\theta})^2}y^2(w_2\phi(y-\theta^*)+w_1\phi(y+\theta^*))\dif y}_{\text{part}~6}.\n
\end{eqnarray}
Since part 5 and part 6 are symmetric with respect to $w_1,w_2$, WLOG, we assume $w_1\geq 0.5$. Then for part 5, note that since $\theta \geq \theta^*$, we have $w_1e^{y\theta^*}+w_2e^{-y\theta^*}\leq w_1e^{y\theta}+w_2e^{-y\theta}$, and therefore,
\begin{eqnarray}
\text{part}~5&\leq&\int_{y\geq 0} \frac{4w_1w_2}{(w_1e^{y\theta^*}+w_2e^{-y\theta^*})^2}y^2(w_1\phi(y-\theta^*)+w_2\phi(y+\theta^*))\dif y\n\\
&=&\int_{y\geq 0} \frac{4w_1w_2}{w_1e^{y\theta^*}+w_2e^{-y\theta^*}}y^2\phi(y)e^{-\frac{(\theta^*)^2}{2}}\dif y\n\\
&\leq&\int_{y\geq 0} 2\sqrt{w_1w_2}y^2\phi(y)e^{-\frac{(\theta^*)^2}{2}}\dif y \kleq \frac{e^{-\frac{(\theta^*)^2}{2}}}{2},\label{eq:Heq_eq10}
\end{eqnarray}
where last two inequalities hold due to AM-GM inequality. For part 6, we have if $\theta\geq \theta^*$,
\begin{eqnarray}
\text{part}~6&=&\int_{y\geq 0} \frac{4}{\left(\sqrt{\frac{w_1}{w_2}}e^{-y\theta}+\sqrt{\frac{w_2}{w_1}}e^{y\theta}\right)^2}y^2(w_1\phi(y+\theta^*)+w_2\phi(y-\theta^*))\dif y\n\\
&\stackrel{(a)}{\leq}&\int_{y\geq 0} \frac{2}{e^{(y-\frac{\ln(w_1/w_2)}{2\theta})\theta}+e^{-(y-\frac{\ln(w_1/w_2)}{2\theta})\theta}}y^2(w_1\phi(y+\theta^*)+w_2\phi(y-\theta^*))\dif y\n\\
&\stackrel{(b)}{\leq}&\int_{y\geq 0}\frac{2}{e^{(y-\frac{\ln(w_1/w_2)}{2\theta})\theta^*}+e^{-(y-\frac{\ln(w_1/w_2)}{2\theta})\theta^*}}y^2(w_1\phi(y+\theta^*)+w_2\phi(y-\theta^*))\dif y\n\\
&=&\int_{y\geq 0} \frac{2}{(\frac{w_2}{w_1})^{\frac{\theta^*}{2\theta}}e^{y\theta^*}+(\frac{w_1}{w_2})^{\frac{\theta^*}{2\theta}}e^{-y\theta^*}}y^2(w_1\phi(y+\theta^*)+w_2\phi(y-\theta^*))\dif y,\label{eq:Heq_eq11}
\end{eqnarray}
where inequality (a) holds due to AM-GM inequality, and inequality (b) holds due to the monotonic of hyperbolic cosine function. Our next step is to prove for all $y\theta^*\geq 0$ and $0< \theta^*\leq \theta$, we have
\begin{eqnarray}
(\frac{w_2}{w_1})^{\frac{\theta^*}{2\theta}}e^{y\theta^*}+(\frac{w_1}{w_2})^{\frac{\theta^*}{2\theta}}e^{-y\theta^*}\geq 2(w_1e^{-y\theta^*}+w_2e^{y\theta^*}),\label{eq:Heq_eq12}
\end{eqnarray}
which, with \eqref{eq:Heq_eq11}, immediately implies that 
$$\text{part}~6  \kleq \int_{y\geq 0} y^2\phi(y)e^{-\frac{(\theta^*)^2}{2}}\dif y \= \frac{e^{-\frac{(\theta^*)^2}{2}}}{2},$$
and therefore, combine with \eqref{eq:Heq_eq10}, we have \eqref{eq:Heq_eq2} holds. To prove \eqref{eq:Heq_eq12}, note that this is equivalent to prove
\begin{eqnarray}
\left(\frac{w_2}{w_1}\right)^{\frac{\theta^*}{2\theta}}\left(1-2w_1^{\frac{\theta^*}{2\theta}}w_2^{1-\frac{\theta^*}{2\theta}}\right)e^{y\theta^*}\geq \left(\frac{w_1}{w_2}\right)^{\frac{\theta^*}{2\theta}}\left(2w_2^{\frac{\theta^*}{2\theta}}w_1^{1-\frac{\theta^*}{2\theta}}-1\right)e^{-y\theta^*}. \label{eq:Heq_eq13}
\end{eqnarray}
Note that
\begin{eqnarray}
w_1^{\frac{\theta^*}{2\theta}}w_2^{1-\frac{\theta^*}{2\theta}}+w_1^{1-\frac{\theta^*}{2\theta}}w_2^{\frac{\theta^*}{2\theta}}&=&(w_1w_2)^{\frac{\theta^*}{2\theta}}(w_1^{1-\frac{\theta^*}{\theta}}+w_2^{1-\frac{\theta^*}{\theta}})\n\\
&\leq&\frac{w_1^{1-\frac{\theta^*}{\theta}}+w_2^{1-\frac{\theta^*}{\theta}}}{2^{\frac{\theta^*}{\theta}}}\n\\
&\leq&(w_1+w_2)^{1-\frac{\theta^*}{\theta}}\ =\ 1.\n
\end{eqnarray}
where the last two inequalities holds due to AM-GM inequality and Holder inequality respectively. Also, since $w_1\geq w_2$, we have 
$$w_1^{\frac{\theta^*}{2\theta}}w_2^{1-\frac{\theta^*}{2\theta}}\leq w_1^{1-\frac{\theta^*}{2\theta}}w_2^{\frac{\theta^*}{2\theta}}.$$
Hence, we have
\begin{eqnarray}
1-2w^{\frac{\theta^*}{2\theta}}w_2^{1-\frac{\theta^*}{2\theta}} \kgeq 0.\n
\end{eqnarray}
Therefore, to prove \eqref{eq:Heq_eq13}, it is sufficient to prove
\begin{eqnarray}
(\frac{w_2}{w_1})^{\frac{\theta^*}{2\theta}}(1-2w_1^{\frac{\theta^*}{2\theta}}w_2^{1-\frac{\theta^*}{2\theta}})&\geq& (\frac{w_1}{w_2})^{\frac{\theta^*}{2\theta}}(2w_2^{\frac{\theta^*}{2\theta}}w_1^{1-\frac{\theta^*}{2\theta}}-1),\n
\end{eqnarray}
which is equivalent to
\begin{eqnarray}
(\frac{w_2}{w_1})^{\frac{\theta^*}{2\theta}}+(\frac{w_1}{w_2})^{\frac{\theta^*}{2\theta}}&\geq& 2(w_1+w_2)=2,\n
\end{eqnarray}
which holds due to AM-GM inequality. Hence, we have \eqref{eq:Heq_eq12} holds. 

\subsection{Proof of Lemma \ref{lem:c2b}}\label{sec:proofc2b}
We first analyze the condition that can determine the sign of $g(\theta,w_1)-w_1$. Note that (with $w_2=1-w_1$)
\begin{eqnarray}
\lefteqn{\frac{g(\theta,w_1)-w_1}{w_1}\=\int \frac{1}{\sqrt{2\pi}}e^{-\frac{y^2+(\theta^*)^2}{2}}\cdot \left(\frac{e^{y\theta}\left(w^*_1e^{y\theta^*}+w^*_2e^{-y\theta^*}\right)}{w_1e^{y\theta}+w_2e^{-y\theta}}-e^{y\theta^*}\right)\dif y} \n\\
&=&\int_{y\geq 0} \frac{1}{\sqrt{2\pi}}e^{-\frac{y^2+(\theta^*)^2}{2}}\cdot \left(\frac{e^{y\theta}\left(w^*_1e^{y\theta^*}+w^*_2e^{-y\theta^*}\right)}{w_1e^{y\theta}+w_2e^{-y\theta}}-e^{y\theta^*}+\frac{e^{-y\theta}\left(w^*_1e^{-y\theta^*}+w^*_2e^{y\theta^*}\right)}{w_1e^{-y\theta}+w_2e^{y\theta}}-e^{-y\theta^*}\right)\dif y \n
\end{eqnarray}
Hence, to determine the sign of $g(\theta,w_1)-w_1\gtrless 0$, we just need to show $\forall y\geq 0$
\begin{eqnarray}
&&\left(\frac{e^{y\theta}\left(w^*_1e^{y\theta^*}+w^*_2e^{-y\theta^*}\right)}{w_1e^{y\theta}+w_2e^{-y\theta}}-e^{y\theta^*}\right) +\left(\frac{e^{-y\theta}\left(w^*_1e^{-y\theta^*}+w^*_2e^{y\theta^*}\right)}{w_1e^{-y\theta}+w_2e^{y\theta}}-e^{-y\theta^*}\right)\gtrless 0,\n
\end{eqnarray}
which is equivalent to
\begin{eqnarray}
(2w_1-1) \cosh_y(\theta^*)+(w^*_1-w_1)\cosh_y(\theta^*+2\theta)+(1-w_1-w^*_1)\cosh_y(\theta^*-2\theta)\gtrless 0, \label{eq:wcp1}
\end{eqnarray}
where $\cosh_y(x)=(e^{yx}+e^{-yx})/2$. Let $\taga=\gamma\theta^*=\frac{2w^*_1-1}{2w_1-1}\theta^*$. Let us first show that for $w_1\in (0.5,1]$
\begin{eqnarray}
g_w(\taga,w^*_1)&\gtrless& w^*_1, \quad \forall w_1\gtrless w_1^*. \label{eq:c2b_eq1}
\end{eqnarray}
By \eqref{eq:wcp1}, we just need to show 
\begin{eqnarray}
\cosh_y(\theta^*)&\gtrless&\cosh_y(\theta^*-2\taga),\quad \forall w_1\gtrless w_1^*, \n
\end{eqnarray}
which holds due to the monotonic of hyperbolic cosine function. Hence, we have proved \eqref{eq:c2b_eq1}. Next, we want to show
\begin{eqnarray}
g_w(\taga,w_1)&\gtrless& w_1, \quad \forall w_1\lessgtr w^*_1. \label{eq:c2b_eq2}
\end{eqnarray}
By \eqref{eq:wcp1}, we just need to show that $\forall y>0$,
\begin{eqnarray}
(2w_1-1) \cosh_y(\theta^*)+(w^*_1-w_1)\cosh_y(\theta^*+2\taga)+(1-w_1-w^*_1)\cosh_y(\theta^*-2\taga)\gtrless 0, \quad \forall w_1\lessgtr w^*_1.\n\\
\label{eq:c2b_eq3}
\end{eqnarray}
Note that, by Taylor expansion of $2\cosh_y(x)=\sum_{i=0}^{\infty}\frac{(xy)^{2i}}{(2i)!}$, we just need to show that given $\gamma=\frac{2w^*_1-1}{2w_1-1}$, we have
\begin{eqnarray}
(2w_1-1)+(w^*_1-w_1)(1+2\gamma)^{2k}+(1-w^*_1-w_1)(2\gamma-1)^{2k}&>&0,\quad  \forall w_1\in (\frac{1}{2},w^*_1), k>0,\n\\
\label{eq:v2}\\
(w_1-w^*_1)(1+2\gamma)^{2k}+(w^*_1+w_1-1)(2\gamma-1)^{2k}-(2w_1-1)&>&0,\quad  \forall w_1\in (w^*_1,1], k>1.\n\\
\label{eq:v1}
\end{eqnarray} 
For \eqref{eq:v2}, since $w_1<w^*_1$, we have $\gamma>1$ and
\begin{eqnarray}
\lefteqn{(2w_1-1)+(w^*_1-w_1)(1+2\gamma)^{2k}+(1-w^*_1-w_1)(2\gamma-1)^{2k}}\n\\
&=&(w^*_1-w_1)\left((1+2\gamma)^{2k}-(2\gamma-1)^{2k}\right)+(2w_1-1)\left(1-(2\gamma-1)^{2k}\right)\n\\
&=&(w^*_1-w_1)\cdot 2\left(\sum_{i=0}^{2k-1}(1+2\gamma)^{i}(2\gamma-1)^{2k-1-i}\right)+(2w_1-1)\cdot (2\gamma-2)\left(\sum_{i=0}^{2k-1}(2\gamma-1)^{i}\right)\n\\
&=&2(w^*_1-w_1)\left(\sum_{i=0}^{2k-1}\left((1+2\gamma)^{i}-2\right)(2\gamma-1)^{2k-1-i}\right)\n\\
&\geq&2(w^*_1-w_1)\left(\sum_{i=0}^{1}\left((1+2\gamma)^{i}-2\right)(2\gamma-1)^{2k-1-i}\right)\n\\
&=&4(w^*_1-w_1)(\gamma-1)(2\gamma-1)^{2k-2}> 0.\n
\end{eqnarray}
For \eqref{eq:v1}, we have
\begin{eqnarray}
\lefteqn{(w_1-w^*_1)(2\gamma+1)^{2k}+(w^*_1+w_1-1)(2\gamma-1)^{2k}-(2w_1-1)}\n\\
&=&(w_1-w^*_1)\left((2\gamma+1)^{2k}-(2\gamma-1)^{2k}\right)+(2w_1-1)\left((2\gamma-1)^{2k}-1\right)\n\\
&=&(w_1-w^*_1)\left((2\gamma+1)^2-(2\gamma-1)^2\right)\left(\sum_{i=0}^{k-1}(2\gamma+1)^{2i}(2\gamma-1)^{2k-2i-2}\right)\n\\
&&+(2w_1-1)\left((2\gamma-1)^{2}-1\right)\left(\sum_{i=0}^{k-1}(2\gamma-1)^{2i}\right)\n\\
&=&8(w_1-w^*_1)\gamma\left(\sum_{i=0}^{k-1}\left((2\gamma+1)^{2i}-1\right)(2\gamma-1)^{2k-2i-2}\right)>0\n.
\end{eqnarray}
Hence, this completes the proof for \eqref{eq:c2b_eq2}.

\subsection{Proof of Lemma \ref{lem:bound}}\label{sec:proofbound}
We just need to bound $\|G_{\theta}(\vtheta,w_1;\vtheta^*,w^*_1)\|^2$. Note that by \eqref{eq:update2theta} and Jensen's inequality, we have
\begin{eqnarray}
\|G_{\theta}(\vtheta,w_1;\vtheta^*,w^*_1)\|^2&\leq&\bbE_{\vy} \left[\left(\frac{w_1e^{\dotp{\vy,\vtheta}}-\pw{t}_2e^{-\dotp{\vy,\vtheta}}}{w_1e^{\dotp{\vy,\theta}}+w_2e^{-\dotp{\vy,\vtheta}}}\right)^2\|\vy\|^2\right]\n\\
&\leq&\bbE_{\vy}\|\vy\|^2 \=1+\|\vtheta^*\|^2.\n
\end{eqnarray}

\subsection{Proof of Lemma \ref{lem:c2c2c}}\label{sec:proofc2c2c}
To show \eqref{eq:v3}, we first define $\taga=\gamma\theta^*$, $\theta_b=b\theta^*$, and
\begin{eqnarray}
A&=&\int y\frac{e^{y\taga}}{w_1e^{y\taga}+(1-w_1)e^{-y\taga}}\left(w^*_1\phi(y-\theta^*)+w^*_2\phi(y+\theta^*)\right)\dif y\n\\
B&=&\int y\frac{e^{-y\taga}}{w_1e^{y\taga}+(1-w_1)e^{-y\taga}}\left(w^*_1\phi(y-\theta^*)+w^*_2\phi(y+\theta^*)\right)\dif y.\n
\end{eqnarray}
Note that $\forall w_1$ 
\begin{eqnarray}
(2w_1-1)\taga&\equiv&w_1A+w_2B.\label{eq:ABrel}
\end{eqnarray}
Hence, we have \eqref{eq:v3} is equivalent to show that
\begin{eqnarray}
w_1A-w_2B&<&\frac{w_1A+w_2B}{2w_1-1}, \quad \forall w_1\in (0.5,w^*_1),\n
\end{eqnarray}
which is equivalent to show
\begin{eqnarray}
A+B&>&0, \quad \forall w_1\in (0.5,w^*_1).\label{eq:c2c2c_eq1}
\end{eqnarray}
Note that 
\begin{eqnarray}
\lefteqn{A+B\=\int \frac{1}{\sqrt{2\pi}}y(e^{y\taga}+e^{-y\taga})e^{-\frac{y^2+(\theta^*)^2}{2}}\frac{w^*_1e^{y\theta^*}+w^*_2e^{-y\theta^*}}{w_1e^{y\taga}+w_2e^{-y\taga}}\dif y}\n\\
&=&\int_{y\geq 0} \frac{1}{\sqrt{2\pi}}y(e^{y\taga}+e^{-y\taga})e^{-\frac{y^2+(\theta^*)^2}{2}}\left(\frac{w^*_1e^{y\theta^*}+w^*_2e^{-y\theta^*}}{w_1e^{y\taga}+w_2e^{-y\taga}}-\frac{w^*_1e^{-y\theta^*}+w^*_2e^{y\theta^*}}{w_1e^{-y\taga}+w_2e^{y\taga}}\right)\dif y\n\\
&=&\int_{y\geq 0} \frac{1}{\sqrt{2\pi}}y(e^{y\taga}+e^{-y\taga})e^{-\frac{y^2+(\theta^*)^2}{2}}\n\\
&&\times \frac{(w^*_1+w_1-1)\left(e^{y\theta^*(1-\gamma)}-e^{-y\theta^*(1-\gamma)}\right)+(w^*_1-w_1)\left(e^{y\theta^*(1+\gamma)}-e^{-y\theta^*(1+\gamma)}\right)}{\left(w_1e^{y\taga}+w_2e^{-y\taga}\right)\left(w_1e^{-y\taga}+w_2e^{y\taga}\right)}\dif y.\n
\end{eqnarray}
Hence, we just need to show that for $\forall y> 0, w^*_1,w_1\in (\frac{1}{2},1)$,
\begin{eqnarray}
(w^*_1+w_1-1)\left(e^{y\theta^*(1-\gamma)}-e^{-y\theta^*(1-\gamma)}\right)+(w^*_1-w_1)\left(e^{y\theta^*(1+\gamma)}-e^{-y\theta^*(1+\gamma)}\right)&>& 0, \quad \forall w_1\in (0.5, w^*_1)\n
\end{eqnarray}
By Taylor expansion of $e^x$, we just need to prove that for all $k\geq 0$, we have
\begin{eqnarray}
(w^*_1+w_1-1)(1-\gamma)^{2k+1}+(w^*_1-w_1)(1+\gamma)^{2k+1}&>&0,\quad  \forall w_1\in (0.5, w^*_1)\,\n
\end{eqnarray}
By definition of $\gamma$, we just need to show
\begin{eqnarray}
(w^*_1+w_1-1)2^{2k+1}(w_1-w^*_1)^{2k+1}+(w^*_1-w_1)2^{2k+1}(w^*_1+w_1-1)^{2k+1}&>&0,\quad \forall  w_1\in (0.5, w^*_1)\,\n\\
\Leftrightarrow w_1+w^*_1-1&>&w^*_1-w_1,\quad \forall w_1\in (0.5, w^*_1),\n
\end{eqnarray}
which obviously holds. To show \eqref{eq:v4}, we should analyze the condition for  $g_{\theta}(\theta,w_1)-\theta>0$. Note that
\begin{eqnarray}
\lefteqn{g_{\theta}(\theta_b,w_1)-\theta_b\=\int y\left(\frac{w_1e^{y\theta_b}-w_2e^{-y\theta_b}}{w_1e^{y\taga}+w_2e^{-y\taga}}-\frac{b}{w^*_1-w^*_2}\right)\left(w^*_1\phi(y-\theta^*)+w^*_2\phi(y+\theta^*)\right)\dif y}\n\\
&=&\frac{1}{w^*_1-w^*_2}\int y\frac{w_1(2w^*_1-1-b)e^{y\theta_b}-w_2(2w^*_1-1+b)e^{-y\theta_b}}{w_1e^{y\taga}+w_2e^{-y\taga}}\left(w^*_1\phi(y-\theta^*)+w^*_2\phi(y+\theta^*)\right)\dif y\n\\
&=&\int_{y\geq 0}\frac{y}{\sqrt{2\pi}}e^{-\frac{y^2+(\theta^*)^2}{2}}\left(\frac{w_1w_2\left((1-b)\cdot 2\sinh_{y\theta^*}(2b+1)+(1+b)\cdot 2\sinh_{y\theta^*}(2b-1)\right)}{\left(w_1e^{y\taga}+w_2e^{-y\taga}\right)\left(w_1e^{-y\taga}+w_2e^{y\taga}\right)}\right.\n\\
&&+\left.\frac{\left((2w_1-1)(2w^*_1-1)-\left(1-2w_1w_2\right)b\right)\cdot 2\sinh_{y\theta^*}(1)}{\left(w_1e^{y\taga}+w_2e^{-y\taga}\right)\left(w_1e^{-y\taga}+w_2e^{y\taga}\right)}\right)\dif y,\n
\end{eqnarray}
where $\sinh_{y\theta^*}(x)=(e^{yx\theta^*}-e^{-yx\theta^*})/2$. Hence, we just need to show for all $y>0$,
\begin{eqnarray}
\lefteqn{w_1w_2\left((1-b)\sinh_{y\theta^*}(2b+1)+(1+b)\sinh_{y\theta^*}(2b-1)\right)}\n\\
&&+\left((2w_1-1)(2w^*_1-1)-\left(1-2w_1w_2\right)b\right)\sinh_{y\theta^*}(1)\ > 0, \quad \forall b\in (0,\gamma],w_1\in (w^*_1,1).\n
\end{eqnarray}
By Taylor expansion of $\sinh_{y\theta^*}(x)$, we just need to show for all $k\geq 0$, we have
\begin{eqnarray}
\lefteqn{w_1w_2\left((1-b)(2b+1)^{2k+1}+(1+b)(2b-1)^{2k+1}\right)}\n\\
&&+\left((2w_1-1)(2w^*_1-1)-\left(1-2w_1w_2\right)b\right)\ \geq 0, \quad \forall b\in (0,\gamma],w_1\in (w^*_1,1).\label{eq:v5}
\end{eqnarray}
where inequality is strict for $k\geq 2$. It is straight forward to check \eqref{eq:v5} holds for $k=0$ due to $b\leq \gamma$. For $k\geq 1$, note that 
\begin{eqnarray}
\lefteqn{(1-b)(2b+1)^{2k+1}+(1+b)(2b-1)^{2k+1}}\n\\
&=&\left((2b+1)^{2k+1}+(2b-1)^{2k+1}\right)-b\left((2b+1)^{2k+1}-(2b-1)^{2k+1}\right)\n\\
&=&4b\sum_{i=0}^{2k}(-1)^{i}(2b+1)^{2k-i}(2b-1)^{i}-2b\sum_{i=0}^{2k}(2b+1)^{2k-i}(2b-1)^{i}\n\\
&=&2b\left(\sum_{i=0}^{k-1}(2b+1)^{2k-2i-1}(2b-1)^{2i}(2b+1-3(2b-1))+(2b-1)^{2k}\right)\n\\
&=&2b+2b\left(\sum_{i=0}^{k-1}(2b+1)^{2k-2i-1}(2b-1)^{2i}(4-4b)+(2b-1)^{2k}-1\right)\n\\
&=&2b+2b(4-4b)\left(\sum_{i=0}^{k-1}(2b+1)^{2k-2i-1}(2b-1)^{2i}-\sum_{i=0}^{k-1}(2b-1)^{2i}b\right)\n\\
&\geq&2b+2b(4-4b)\left(\sum_{i=0}^{k-1}(b+1)(2b-1)^{2i}\right)\n\\
&\geq&2b.\n
\end{eqnarray}
where last two inequalities hold due to $b\leq \gamma<1$ and last inequality is strict when $k\geq 2$. Hence, to show \eqref{eq:v5}, we just need to show
\begin{eqnarray}
&&2bw_1w_2+(2w_1-1)(2w^*_1-1)-\left(1-2w_1w_2\right)b\ \geq \ 0\n\\
&\Leftrightarrow& b\ \leq \ \gamma,\n
\end{eqnarray}
which holds clearly. Hence, this completes the proof for this lemma.

\section{Additional numerical results}\label{sec:tables}
\begin{table}[h]
\begin{center}
\begin{tabular}{|c|c|c|c|c|}
  \hline
  Sample size &
  Separation &
  $w^*_1=0.52$ & $w^*_1=0.7$ & $w^*_1=0.9$  \\
  \hline 
  \multirow{3}*{$n=1000$} & $\theta^*_2=1$ & \spacefrac{\textcolor{red}{0.999}}{\textcolor{blue}{0.999}} & \spacefrac{\textcolor{red}{0.499}}{\textcolor{blue}{0.699}} & \spacefrac{\textcolor{red}{0.450}}{\textcolor{blue}{0.338}} \\
  \cline{2-5}
  & $\theta^*_2=2$ & \spacefrac{\textcolor{red}{0.799}}{\textcolor{blue}{0.500}} & \spacefrac{\textcolor{red}{0.497}}{\textcolor{blue}{0.800}} & \spacefrac{\textcolor{red}{0.499}}{\textcolor{blue}{0.899}}\\
  \cline{2-5}
     & $\theta^*_2=4$ & \spacefrac{\textcolor{red}{1.000}}{\textcolor{blue}{1.000}} & \spacefrac{\textcolor{red}{0.447}}{\textcolor{blue}{0.900}} & \spacefrac{\textcolor{red}{0.501}}{\textcolor{blue}{0.999}} \\
  \hline
  \multirow{3}*{$n=\infty$} & $\theta^*_2=1$ & \spacefrac{\textcolor{red}{0.497}}{\textcolor{blue}{1.000}} & \spacefrac{\textcolor{red}{0.493}}{\textcolor{blue}{1.000}} & \spacefrac{\textcolor{red}{0.501}}{\textcolor{blue}{0.000}} \\
  \cline{2-5}
    & $\theta^*_2=2$ &\spacefrac{\textcolor{red}{0.504}}{\textcolor{blue}{1.000}} & \spacefrac{\textcolor{red}{0.514}}{\textcolor{blue}{1.000}} & \spacefrac{\textcolor{red}{0.506}}{\textcolor{blue}{1.000}}\\
  \cline{2-5}
     & $\theta^*_2=4$ & \spacefrac{\textcolor{red}{0.495}}{\textcolor{blue}{1.000}} & \spacefrac{\textcolor{red}{0.490}}{\textcolor{blue}{1.000}} & \spacefrac{\textcolor{red}{0.514}}{\textcolor{blue}{1.000}} \\
  \hline
\end{tabular}
\end{center}
\caption{In this table, we consider mixture of two Gaussian in one dimension with $\theta^*_1=0$. We present the probability of success $P_1$ and $P_2$ for EM to find the MLE for Model 1 and Model 2, respectively, reported as
  $\spacefrac{\textcolor{red}{P_1}}{\textcolor{blue}{P_2}}$.
  We only keep the first 3 digits after the decimal for each probability.}
\label{tab:rantab1}
\end{table}

\begin{table}[h]
\begin{center}
\begin{tabular}{|c|c||c|c|c|}
\hline
Sample size      &  Separation & $w^*_1=0.52$ & $w^*_1=0.7$ & $w^*_1=0.9$  \\
\hline 
\hline 
\multirow{3}*{$n=1000$} & $\theta^*_2=1$ & 0.999 & 0.999 & 0.800 \\
\cline{2-5}
& $\theta^*_2=2$ &1.000 &1.000 &1.000\\
\cline{2-5}
   & $\theta^*_2=4$ & 1.000 & 1.000&1.000 \\
\hline
\multirow{3}*{$n=\infty$} & $\theta^*_2=1$ &1.000&1.000 & 1.000 \\
\cline{2-5}
  & $\theta^*_2=2$ &1.000 & 1.000 & 1.000\\
\cline{2-5}
   & $\theta^*_2=4$ & 1.000 &1.000& 1.000\\
\hline
\end{tabular}\\
\vspace{0.2cm}
\begin{tabular}{|c|c|c|c|}
\hline
 Case 1 & Case 2 & Case 3  &Case 4 \\
\hline 
\hline 
0.980 &0.998 & 1.000 &1.000\\
\hline
\end{tabular}
\end{center}
\caption{We present the probabilities of success $P_3$ for EM to find the MLE for Model 1 under the new procedure described in Section~\ref{sec:3G}. The first table is for mixture of two Gaussians in one dimension discussed in Section~\ref{sec:twoG}. The second table is for mixture of three or four Gaussians discussed in Section~\ref{sec:3G}. We only keep the first 3 digits after the decimal for each probability.}
\label{tab:tildeP}
\end{table}

\end{document}